\documentclass[10pt,twocolumn,letterpaper]{article}

\usepackage[pagenumbers]{cvpr} % To force page numbers, e.g. for an arXiv version

\usepackage{booktabs}
\usepackage{caption}
\usepackage{amsmath,amsfonts,bm}

\def\eqref#1{equation~\ref{#1}}
\def\1{\bm{1}}

\def\rmY{{\mathbf{Y}}}

\def\vzero{{\bm{0}}}
\def\vone{{\bm{1}}}
\def\vmu{{\bm{\mu}}}

\def\va{{\bm{a}}}
\def\vb{{\bm{b}}}
\def\vc{{\bm{c}}}
\def\vd{{\bm{d}}}
\def\ve{{\bm{e}}}
\def\vf{{\bm{f}}}

\def\vh{{\bm{h}}}

\def\vq{{\bm{q}}}

\def\vs{{\bm{s}}}
\def\vt{{\bm{t}}}
\def\vu{{\bm{u}}}
\def\vv{{\bm{v}}}

\def\vx{{\bm{x}}}

\def\vz{{\bm{z}}}
\def\veps{{\bm{\epsilon}}}
\def\vdelta{{\bm{\delta}}}
\def\mPi{{\bm{\Pi}}}

\def\mC{{\bm{C}}}

\def\mG{{\bm{G}}}

\def\mI{{\bm{I}}}

\def\mK{{\bm{K}}}
\def\mL{{\bm{L}}}

\def\mP{{\bm{P}}}

\def\mR{{\bm{R}}}

\def\mU{{\bm{U}}}
\def\mV{{\bm{V}}}

\def\mPhi{{\bm{\Phi}}}

\def\mSigma{{\bm{\Sigma}}}

\DeclareMathAlphabet{\mathsfit}{\encodingdefault}{\sfdefault}{m}{sl}
\SetMathAlphabet{\mathsfit}{bold}{\encodingdefault}{\sfdefault}{bx}{n}

\def\gD{{\mathcal{D}}}
\def\gE{{\mathcal{E}}}

\def\gL{{\mathcal{L}}}

\def\gN{{\mathcal{N}}}

\def\gS{{\mathcal{S}}}

\def\gY{{\mathcal{Y}}}

\def\sR{{\mathbb{R}}}

\def\emK{{K}}
\def\emL{{L}}

\newcommand{\E}{\mathbb{E}}

\newcommand{\KL}{D_{\mathrm{KL}}}

\DeclareMathOperator*{\argmax}{arg\,max}

\DeclareMathOperator{\Tr}{Tr}

\definecolor{cvprblue}{rgb}{0.21,0.49,0.74}
\definecolor{oursbg}{RGB}{230,240,255}
\usepackage[pagebackref,breaklinks,colorlinks,allcolors=cvprblue]{hyperref}
\usepackage{algorithm}
\usepackage{algpseudocode}
\usepackage{amsthm, amssymb}
\usepackage{thmtools,thm-restate}
\usepackage{multirow}
\usepackage{graphicx}
\usepackage[table]{xcolor} 
\usepackage{caption}

\def\confName{CVPR}
\def\confYear{2027}

\title{TWIST: Teacher-Weighted Initial-Noise Spherical Transform for Quality-Preserving Diversity in Distilled Diffusion Models}

\author{Shayan Baghayi-Nejad \qquad Jeongsol Kim \qquad Jong Chul Ye\\
KAIST\\
{\tt\small \{shayan,jeongsol,jong.ye\}@kaist.ac.kr}
}

\declaretheorem[name=Lemma,       refname={Lemma,Lemmas},            Refname={Lemma,Lemmas}]{lemma}

\begin{document}
% \crefname{prop}{Proposition}{Propositions}
% \Crefname{prop}{Proposition}{Propositions}

\twocolumn[{%
\renewcommand\twocolumn[1][]{#1}%
\maketitle
\begin{center}
  \centering
  \captionsetup{type=figure}
  \includegraphics[width=\textwidth]{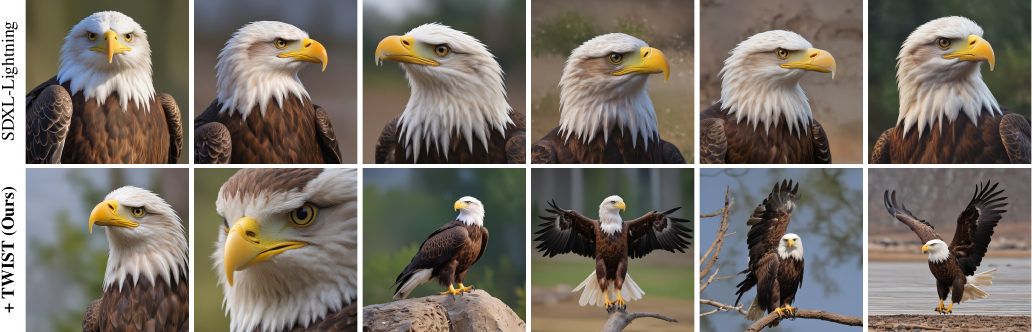}
  \captionof{figure}{\textbf{TWIST restores the diversity lost in few-step distillation.}
  Samples for the prompt ``a photo of a bald eagle''.
  \emph{Top:} SDXL-Lightning (4 steps) collapses to near-identical head portraits.
  \emph{Bottom:} the same frozen model with TWIST, which only changes the input noise, produces varied poses, scales and scenes while maintaining image quality. Both subsets are chosen from a set of 32 images with the objective of maximizing a quality-weighted Determinantal Point Processes (DPP) determinant.}
  \label{fig:teaser}
\end{center}%
}]

% =============================
% =         ABSTRACT          =
% =============================

\begin{abstract}

Timestep-distilled diffusion models generate high-quality images in one or a few steps, but often suffer from reduced mode coverage and limited diversity.
Recent methods improve diversity by optimizing the initial noise at inference time, requiring costly backpropagation through the denoiser.
Moreover, their diversity objectives typically repel generated samples without accounting for how the corresponding teacher model distributes its outputs.
We introduce \textbf{T}eacher-\textbf{W}eighted \textbf{I}nitial-noise \textbf{S}pherical \textbf{T}ransform (TWIST), a lightweight network that transforms the low-frequency components of initial noise to improve diversity without inference-time optimization.
TWIST uses a teacher-guided determinantal point process (DPP) objective to align the feature-space diversity of student samples with that of the teacher.
By amortizing noise optimization into a single network, TWIST eliminates inference-time backpropagation while preserving the efficiency of few-step generation.
On SDXL-Turbo, TWIST matches the diversity of MoNO at less than 25\% of its inference cost, while achieving higher recall and better PickScore, ImageReward, and CLIPScore under matched compute budgets.
On SDXL-Lightning, it achieves comparable recall at lower cost without the ImageReward degradation observed with MoNO.
\end{abstract}

% =============================
% =       INTRODUCTION        =
% =============================

% \input{sec/1_intro}
\section{Introduction}
\label{sec:intro}
Diffusion and flow models~\cite{ho2020ddpm,song2021sde,rombach2022ldm, lipman2022flow} have become one of the main paradigms for conditional text-to-image (T2I) generation. However, they require many denoising steps to produce a photorealistic output. Distillation techniques~\cite{salimans2022progressive,song2023consistency,sauer2023add,yin2024dmd,yin2024dmd2,lin2024lightning} enable fast T2I generation in one or a few steps by learning to take larger steps toward the data manifold. This high-fidelity few-step generation comes at the cost of severe diversity collapse~\cite{gandikota2025distilling}, where generations from different noises look very similar.

This limitation of few-step generators has motivated methods that increase the diversity of distilled diffusion models without losing quality. Recent works take several directions. One line of work perturbs the conditioning or the guidance~\cite{sadat2024cads,kynkaanniemi2024interval}, manipulating conditional signals or intermediate feature representations to avoid collapse to specific modes. Another family generates many candidates and selects the subset with the best diversity-quality trade-off~\cite{parmar2025gi}. Most recently, noise optimization has emerged as a promising direction for diversifying the outputs~\cite{kim2025cno,harrington2026divgen,shi2026mono}.

Guidance perturbation and noise optimization methods rely either on random noise or on repelling samples from each other in a feature space. While repulsion increases diversity, nothing guarantees that the pushed samples move in a meaningful direction or stay aligned with the prompt. These methods therefore have no notion of quality when steering the noise, and have to rely on anchors, rewards and regularizers to maintain it, which may conflict with the diversity objective and make the optimization unstable. Moreover, inference-time noise optimization requires backpropagation through the denoiser and additional forward passes for every generation, which is expensive compared to one-step generation.

In this work, we go beyond batch repulsion and volume maximization by showing the model we optimize what the spread of a batch of high-quality, diverse samples should look like. Specifically, we propose \textit{Teacher-Weighted Initial-noise Spherical Transform (TWIST)}, a teacher-guided framework in which a small convolutional U-Net is trained on the reference model's samples to rotate the initial noise so that the shape of the student's batch in a feature space approaches that of the teacher's. TWIST rotates only the low-frequency component of the initial noise, on the sphere of Gaussian noise, before passing it to the denoiser. 

Our objective builds on {\em Determinantal Point Processes (DPP)}~\cite{kulesza2012dpp}, where the eigen-decomposition of the kernel built from the samples' features describes how the samples span that space. Inspired by GDPP~\cite{elfeki2019gdpp}, we match both the eigenvalues and their corresponding eigenvectors between the teacher's and the student's kernels. The eigenvalues set how much the samples should spread, and the eigenvectors set the directions in which they should spread. The volume is therefore not expanded naively, and since the teacher's directions carry quality information, we need no reward model or anchor to the original sample; the only constraints are on the noise itself, which stays on the Gaussian sphere with a bounded rotation and a KL term that keeps it close to $\gN(\mathbf{0}, \mI)$. 
TWIST learns the rotation in a training phase and adds only a small U-Net forward pass at inference, which keeps generation fast. Because the generator needs no backpropagation, any additional budget can be spent on forward passes alone.

To evaluate TWIST, we compare it against inference-time noise optimization at the same cost through a quality-weighted DPP selection over candidates generated by TWIST.
As shown in \Cref{fig:teaser}, TWIST restores diversity in SDXL-Lightning, producing varied poses, scales, and backgrounds where the original student generates nearly identical portraits.
A 2D toy experiment further illustrates the advantage of teacher-guided diversity (\cref{fig:toy}).
When the student covers only three of eight modes, repulsion-based methods (CNO, MoNO, and DPP volume maximization) either fail to recover missing modes or generate off-distribution samples, whereas our objective recovers all eight.
Our contributions are as follows:

\begin{itemize}
\item We introduce \textit{reference-guided} initial noise generation, in which naive volume expansion is replaced by expanding the volume toward directions that carry semantics, approximated by teacher-generated samples. We show that matching the eigenvectors, and not only the eigenvalues, is what preserves quality.

\item We train a lightweight noise generator that rotates the initial noise on the Gaussian noise sphere to increase diversity. It costs one small U-Net forward pass before denoising and has \textit{negligible overhead at inference}.

\item On SDXL-Turbo, SDXL-DMD2 and SDXL-Lightning, our method {matches or improves on the diversity-quality trade-off of state-of-the-art noise optimization at equal or lower cost}.
\end{itemize}

\begin{figure*}[t]
  \centering
  \includegraphics[width=0.9\textwidth]{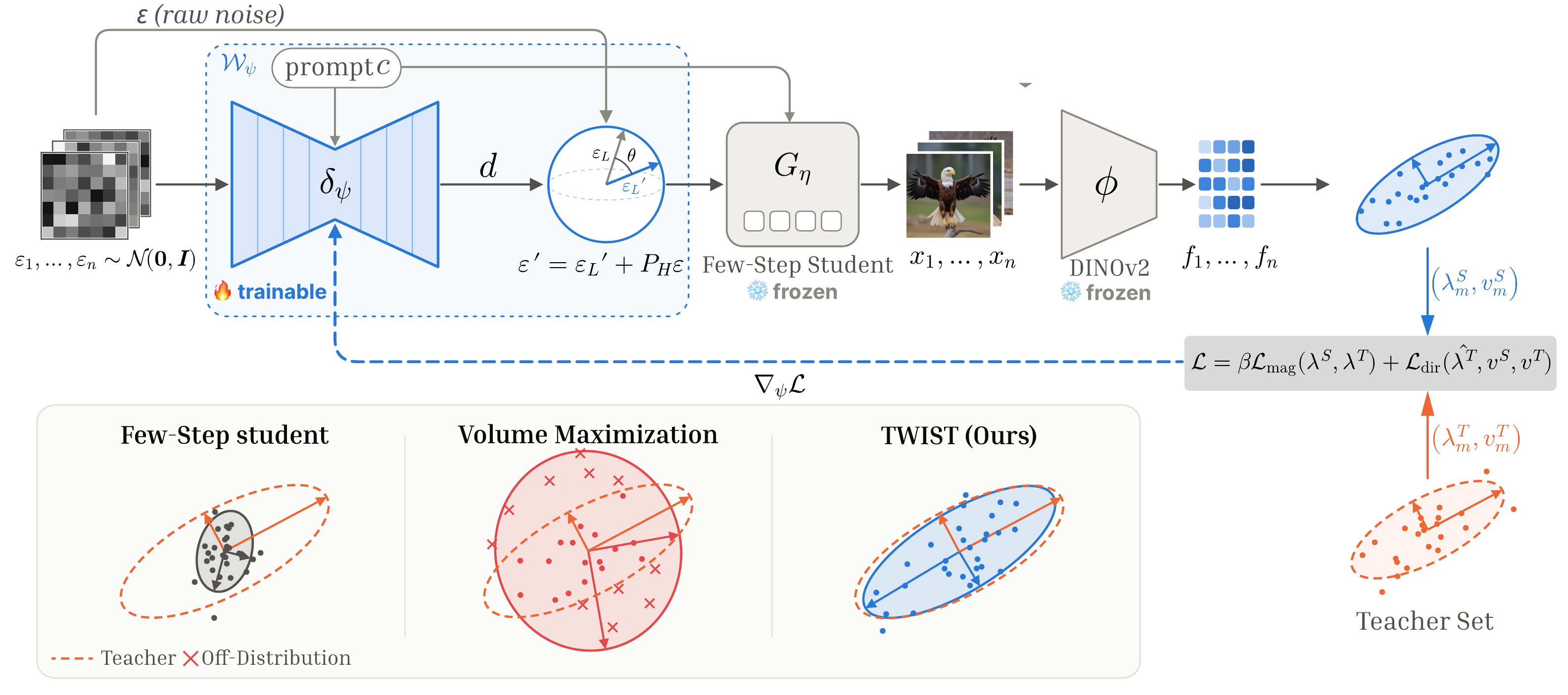}
  \vspace{-0.3cm}
  \caption{\textbf{Overview of TWIST.}
  \emph{Top:} the trainable transform $\mathcal{W}_\psi$ maps a set of input noises $\veps_1, \dots, \veps_n$ to $\veps'_1, \dots, \veps'_n$. The proposal network $\delta_\psi$ outputs a direction $\vd$, which rotates the low-frequency part $\veps_L$ of each noise on its sphere by an angle $\theta$, while the high-frequency part $\mP_H \veps$ is passed through unchanged (\cref{eq:rotate}). The frozen few-step student $G_\eta$ generates images from the transformed noises, and a frozen encoder $\phi$ embeds them. The eigenvalues and eigenvectors of the student set are matched to those of a precomputed teacher set for the same prompt (\cref{eq:twistloss}), and the gradient updates only $\psi$. The KL regularizer of \cref{eq:total} is omitted for clarity.
  \emph{Bottom:} a few-step student covers only part of the teacher's distribution. Maximizing volume alone spreads samples in all directions, including outside the teacher's distribution. TWIST spreads them along the teacher's principal directions, by the teacher's amounts.}
  \label{fig:method}
  \vspace{-0.5cm}
\end{figure*}

% =============================
% =       RELATED WORK        =
% =============================
\section{Related Work}

\noindent\textbf{Diversity collapse in distilled diffusion models.}
Distillation techniques have enabled efficient T2I generation. However, the gain in speed causes the model to shrink into narrow modes and generate nearly identical images under the same conditioning~\cite{gandikota2025distilling}. This has drawn attention to increasing the diversity of few-step generators. Diversity distillation~\cite{gandikota2025distilling} runs the first denoising step with the base model and the remaining steps with the student, recovering diversity at the price of an extra teacher forward pass and some loss in quality.
%In contrast, TWIST learns to recover diversity through a lightweight transformation of the initial noise, using the teacher only during training and preserving few-step generation at inference.

\noindent\textbf{Inference-time diversity optimization.}
Recent methods have focused on training-free diversity optimization at inference. Guidance perturbation methods inject noise into the conditioning during generation~\cite{sadat2024cads} or apply Classifier-Free Guidance (CFG)~\cite{ho2022cfg} only in selected intervals~\cite{kynkaanniemi2024interval} to broaden the distribution of samples. Sampling-based methods use interactions between samples during generation to repel them from each other~\cite{corso2024particle}, or generate many candidates and select a diverse, prompt-aligned subset~\cite{parmar2025gi}. Several of these methods need the denoising trajectory across timesteps~\cite{sadat2024cads,kynkaanniemi2024interval,corso2024particle}, which does not exist in one-step generation. More recently, gradient-based noise optimization has gained popularity. CNO~\cite{kim2025cno} uses a contrastive objective in the downsampled Tweedie prediction space: the samples of a batch repel each other, while each is attracted to the prediction from its original, non-optimized noise, which acts as a fixed anchor to preserve fidelity. DivGen~\cite{harrington2026divgen} maximizes the DPP log-determinant of the batch's features, together with a reward model to keep the quality and a regularizer on the initial noise to avoid artifacts. MoNO~\cite{shi2026mono} removes the need for an anchor and a regularizer by applying Riemannian updates that keep the noise on its original manifold. Its objective minimizes the projection of each new sample onto the space spanned by the previous ones, so that every generation follows an unseen direction. However, this objective requires sequential generation and cannot process the noises of a batch in parallel.
In contrast, TWIST learns a teacher-guided noise transformation that can be applied in parallel, requiring neither denoiser backpropagation nor sequential optimization at inference.

\noindent\textbf{Joint quality-diversity optimization.}
Existing noise optimization methods either add separate loss terms that act as soft quality regularizers, or constrain the noise manifold to avoid artifacts and unrealistic images. Quality terms that pull against the diversity term can flatten or destabilize the learning signal. GDPP~\cite{elfeki2019gdpp} uses determinantal point processes as an additional loss for training Generative Adversarial Networks (GANs) that penalizes the mismatch between the shape of real and generated samples in a feature space. This objective optimizes quality and diversity together: besides stretching the volume spanned by the generated samples, it requires them to take the same shape as the real data instead of growing the volume arbitrarily.
%We adopt this view for distilled diffusion models, where the real data of a prompt is replaced by samples of the multi-step teacher.

Inspired by GDPP, TWIST uses samples from a multi-step teacher as a prompt-specific reference for distilled diffusion models.
By matching both eigenvalues and eigenvectors of the teacher and student kernels, TWIST learns to increase diversity toward the teacher's sample geometry, rather than expanding it indiscriminately.
This integrates quality and diversity into a single teacher-guided objective without requiring a separate reward model or sample anchor.

\section{Preliminaries}
\label{sec:prelim_dpp}

\noindent\textbf{Kernel and volume.}
Let $\phi$ be a feature extractor with $\|\phi(\vx)\|_2 = 1$, and let $\mPhi = [\vf_1, \dots, \vf_n]^\top \in \sR^{n \times d}$ stack the features $\vf_i = \phi(\vx_i)$ of a set of $n$ items. A determinantal point process (DPP)~\cite{kulesza2012dpp} with kernel $\mL = \mPhi\mPhi^\top$ selects a subset $\gS$ with probability $P_{\mL}(\gS) \propto \det(\mL_{\gS})$, the squared volume spanned by $\{\vf_i\}_{i \in \gS}$, so similar items are unlikely to be selected together. With per-item qualities $q_i > 0$, the kernel $\emL_{ij} = q_i \vf_i^\top \vf_j\, q_j$ gives $\det(\mL_{\gS}) = \prod_{i \in \gS} q_i^2 \, \det(\mPhi_{\gS}\mPhi_{\gS}^\top)$, which favors subsets that are both of high quality and spread out. The most likely subset is found greedily~\cite{chen2018fastgreedy}, each step adding
\begin{equation}
  j^\star = \argmax_{j \notin \gS} \; \Big[ \log q_j^2 + \log \big\| (\mI - \mP_{\gS}) \vf_j \big\|_2^2 \Big],
  \label{eq:greedy}
\end{equation}
where $\mP_{\gS}$ projects onto $\mathrm{span}\{\vf_i\}_{i \in \gS}$. MoNO~\cite{shi2026mono} minimizes the projection of each new sample's features onto the span of the previous ones, i.e., it maximizes the second term of \cref{eq:greedy} by gradient descent on the noise.

\noindent\textbf{Spectrum.}
Let $\mL = \sum_{m} \lambda_m \vu_m \vu_m^\top$ with $\lambda_1 \ge \dots \ge \lambda_n \ge 0$. The normalizer of the DPP gives the diversity score of the set,
\begin{equation}
  \gD(\mL) = \log \det(\mL + \mI) = \textstyle\sum_{m=1}^{n} \log(1 + \lambda_m),
  \label{eq:dppscore}
\end{equation}
the log of the total squared volume of all its subsets, which volume-based objectives maximize~\cite{harrington2026divgen}. It depends on the eigenvalues alone: they set how strongly each direction is represented, while the eigenvectors set which directions these are.

\noindent\textbf{Gram--covariance duality.}
With the thin SVD $\mPhi = \mU\mSigma\mV^\top$, the kernel and the feature-space second-moment matrix $\mC = \tfrac{1}{n}\mPhi^\top\mPhi$ satisfy
\begin{equation}
  \mL = \mU\mSigma^2\mU^\top \in \sR^{n \times n}, \quad
  \mC = \tfrac{1}{n}\mV\mSigma^2\mV^\top \in \sR^{d \times d}.
  \label{eq:duality}
\end{equation}
They share their nonzero spectrum, $\lambda_m(\mL) = n\,\lambda_m(\mC) = \sigma_m^2$, and their eigenvectors are related by $\vv_m = \mPhi^\top\vu_m / \sigma_m$. The $\vu_m \in \sR^n$ weight the samples, while the $\vv_m \in \sR^d$ are the directions along which the set varies in feature space. For any orthogonal $\mR \in \sR^{d \times d}$, the rotated set $\mPhi\mR$ has the same kernel $\mL$ but second-moment matrix $\mR^\top\mC\mR$: same eigenvalues, rotated eigenvectors. An objective that depends only on the pairwise similarities $\vf_i^\top\vf_j$ of the generated samples, including \cref{eq:dppscore}, therefore cannot distinguish a set that spreads along the directions in which a reference set varies from one that spreads by the same amount along any other directions.

\noindent\textbf{GDPP.}
GDPP~\cite{elfeki2019gdpp} trains GANs to imitate the diversity of the data rather than merely increase it. From the kernels of a real and a generated batch, with eigen-pairs $(\lambda^R_m, \vu^R_m)$ and $(\lambda^G_m, \vu^G_m)$, it adds to the generator loss
\begin{equation}
  \gL_{\mathrm{GDPP}} = \sum_{m} \big(\lambda^R_m - \lambda^G_m\big)^2
  - \sum_{m} \hat{\lambda}^R_m \cos\big(\vu^R_m, \vu^G_m\big),
  \label{eq:gdpp}
\end{equation}
where $\hat{\lambda}^R_m$ are the min-max normalized real eigenvalues. The first term asks for the same amount of spread as the data and the second for the same directions; by the duality above, only the second fixes the orientation. Its eigenvectors, however, live in sample space, so comparing $\vu^R_m$ with $\vu^G_m$ implicitly treats the $i$-th real and the $i$-th generated sample as corresponding. The feature-space eigenvectors $\vv_m$ of \cref{eq:duality} carry the same structure without this assumption (\cref{sec:objective}).

\section{Method: TWIST}
\label{sec:method}

%\subsection{Overview}
Let $\mathcal{T}(\vz, \vc)$ denote the multi-step teacher sampler and $G_\eta(\vz, \vc)$ the frozen few-step student, both mapping a noise $\vz \sim \gN(\vzero, \mI)$ and a prompt $\vc$ to an image. TWIST places a small trainable transform $\mathcal{W}_\psi$ in front of the frozen student (\cref{fig:method}). For a prompt $\vc$ and a set of input noises $\gE = \{\veps_i\}_{i=1}^{n}$, $\veps_i \sim \gN(\vzero, \mI)$, it produces
\begin{equation}
  \vx_i = G_\eta\big(\veps'_i, \vc\big), \qquad
  \veps'_i = \mathcal{W}_\psi(\veps_i;\, \gE, \vc),
  \label{eq:twist}
\end{equation}
and is trained so that the set $\{\vx_i\}_{i=1}^{n}$ has the same spectral shape as a set of teacher samples of the same prompt. We first describe the transform (\cref{sec:transform}), then the training objective (\cref{sec:objective}), and finally how extra inference budget is spent on forward passes alone (\cref{sec:select}). Architectural and optimization details, and the values of all hyperparameters, are given in the appendix.

\subsection{Spherical Low-Frequency Transform}
\label{sec:transform}

\noindent
\textbf{Proposal network.}
A small convolutional U-Net $\delta_\psi$ operates on the latent grid of the noise. It is conditioned on the prompt's pooled text embedding $\ve_{\vc}$ through FiLM~\cite{perez2018film} and outputs a proposal $\vd_i = \delta_\psi(\veps_i, \ve_{\vc})$ with the shape of the noise. Its last layer is zero-initialized, so training starts from the unmodified student.

\noindent
\textbf{Removing shared components.}
A proposal that moves every noise of the set in the same way shifts the whole set instead of spreading it, and a spatially constant change of a latent channel changes global color or brightness. We remove both by subtracting from each proposal its per-channel spatial mean and then the mean over the set, which gives $\bar{\vd}_i$. The set mean only involves the proposals of $\delta_\psi$, not the student; we discuss how it is formed when images are generated one at a time in \cref{app:additional}.

\noindent
\textbf{Low-frequency split.}
Low frequencies of the initial noise set image layout and are generated
first~\cite{falck2025fourier,harrington2026divgen}, and noise optimization
acts mainly on them~\cite{harrington2026divgen}. Like MoNO~\cite{shi2026mono}
and LENS~\cite{jeon2026lens}, we only change the low band. 
% Let $\mathcal{F}$ be the per-channel 2-D Fourier transform and $(\omega_x, \omega_y)$ the frequencies in cycles per latent pixel. The low-pass projector with cut-off $\rho$ is
% \begin{equation}
%   \mP_L \vx = \mathcal{F}^{-1}\big(\mM_\rho \odot \mathcal{F}\vx\big), \qquad
%   [\mM_\rho]_{\omega} = \mathbb{I}\Big[\sqrt{2(\omega_x^2 + \omega_y^2)} \le \rho\Big].
%   \label{eq:lowpass}
% \end{equation}
% Since the mask is binary and symmetric, $\mP_L$ is an orthogonal projector, and so is $\mP_H = \mI - \mP_L$.
The low-pass orthogonal projector is denoted as $\mP_L$, and its high-pass counterpart is $\mP_H=\mI-\mP_L$.

\noindent
\noindent
\textbf{Rotation on the low-frequency sphere.}
Let $\veps_L = \mP_L \veps$, $\vd_L = \mP_L \bar{\vd}$ (sample index dropped), and $\vt$ the part of $\vd_L$ tangent to the sphere through $\veps_L$. We rotate $\veps_L$ along $\vt$ by an angle with a soft cap $\theta_{\max}$:
\begin{equation}
\begin{aligned}
  \veps'_L &= \cos\theta\, \veps_L + \sin\theta\, \|\veps_L\| \tfrac{\vt}{\|\vt\|}, \\
  \theta &= \theta_{\max} \tanh\!\tfrac{\|\vt\|}{\theta_{\max}\|\veps_L\|}, \\
  \veps' &= \veps'_L + \mP_H \veps.
\end{aligned}
\label{eq:rotate}
\end{equation}
Since $\vt \perp \veps_L$ and the two bands are orthogonal, the rotation keeps $\|\veps'_L\| = \|\veps_L\|$, leaves the high band unchanged, and preserves $\|\veps'\| = \|\veps\|$, and with it the likelihood of the noise under $\gN(\vzero, \mI)$. The norm does not determine the whole distribution of $\veps'$, so we add a light regularizer on its per-channel statistics,
\begin{equation}
  \gL_{\mathrm{KL}} = \frac{1}{nC} \sum_{i=1}^{n} \sum_{c=1}^{C} \KL\big(\gN(m_{ic}, s^2_{ic}) \,\big\|\, \gN(0, 1)\big),
  \label{eq:kl}
\end{equation}
where $m_{ic}$ and $s^2_{ic}$ are the spatial mean and variance of channel $c$ of $\veps'_i$. At inference, \cref{eq:twist} costs one forward pass of $\delta_\psi$ and two FFTs per noise.

\subsection{Teacher-Guided Spectral Objective}
\label{sec:objective}

\noindent
\textbf{Per-prompt sets.}
For every training prompt we pre-compute $n$ teacher samples $\vx^T_i = \mathcal{T}(\vz'_i, \vc)$ and store their noises. The student receives the same noises, $\veps_i = \vz'_i$, average-pooled and rescaled to stay exactly standard normal when its latent is smaller than the teacher's. We embed both sets with a frozen feature extractor $\phi$, the DINOv2~\cite{oquab2024dinov2} patch tokens concatenated into one $\ell_2$-normalized vector per image, which gives $\mPhi_S, \mPhi_T \in \sR^{n \times d}$.

\noindent
\textbf{A common center.}
We center both sets at the teacher's mean $\vmu_T = \frac{1}{n}\mPhi_T^\top \vone$, $\tilde{\mPhi}_S = \mPhi_S - \vone\vmu_T^\top$ and $\tilde{\mPhi}_T = \mPhi_T - \vone\vmu_T^\top$. The teacher's second-moment matrix is then its covariance $\mSigma_T$, while the student's is
\begin{equation}
  \tfrac{1}{n} \tilde{\mPhi}_S^\top \tilde{\mPhi}_S = \mSigma_S + \vdelta\vdelta^\top, \qquad
  \vdelta = \vmu_S - \vmu_T .
  \label{eq:centre}
\end{equation}
A shift of the student's set away from the teacher's thus appears as an additional eigen-direction with no teacher counterpart, and the objective below penalizes it. Centering each set at its own mean would hide the shift; not centering at all would let the mean dominate both spectra.

\noindent
\textbf{Feature-space eigen-pairs.}
Because $d \gg n$, we never form the $d \times d$ matrices. Using the duality of \cref{eq:duality}, we decompose the $n \times n$ Gram matrices $\tilde{\mPhi}\tilde{\mPhi}^\top = \sum_m \lambda_m \vu_m \vu_m^\top$, with eigenvalues in descending order, and map the eigenvectors to feature space, $\vv_m = \tilde{\mPhi}^\top \vu_m / \sqrt{\lambda_m}$. This gives the eigen-pairs $(\lambda^S_m, \vv^S_m)$ and $(\lambda^T_m, \vv^T_m)$ at a cost of $O(n^2 d + n^3)$.

\noindent
\textbf{Objective.}
We match the two spectra in magnitude and in direction:
\begin{equation}
\begin{aligned}
  \gL_{\mathrm{mag}} = \frac{1}{n} \sum_{m=1}^{n} \big(\lambda^S_m - \lambda^T_m\big)^2, 
  \gL_{\mathrm{dir}} = -\sum_{m=1}^{n} \hat{\lambda}^T_m \,\big|\langle \vv^S_m, \vv^T_m \rangle\big|,
\end{aligned}
\label{eq:twistloss}
\end{equation}
where $\hat{\lambda}^T_m = \frac{\lambda^T_m - \min_k \lambda^T_k}{\max_k \lambda^T_k - \min_k \lambda^T_k}$, and train $\psi$ with
\begin{equation}
  \gL(\psi) = \E_{\vc}\Big[\beta\, \gL_{\mathrm{mag}} + \gL_{\mathrm{dir}} + \gamma\, \gL_{\mathrm{KL}}\Big].
  \label{eq:total}
\end{equation}
Gradients flow through $\phi$, the decoder and the frozen student into $\psi$; the teacher's eigen-pairs are constants. 

\Cref{eq:twistloss} is the GDPP loss of \cref{eq:gdpp} with four changes: the eigenvectors are taken in feature space, both sets share the teacher's center (\cref{eq:centre}), the absolute value removes the sign ambiguity of eigenvectors, and the reference is a per-prompt set of teacher samples rather than a batch of real data. Unlike the volume score of \cref{eq:dppscore}, $\gL_{\mathrm{mag}}$ is two-sided: it also penalizes a student that spreads more than the teacher along a direction. $\gL_{\mathrm{dir}}$ pins the orientation that, by the rotation argument of \cref{sec:prelim_dpp}, no similarity-only objective can see. The weights $\hat{\lambda}^T_m$ concentrate the direction term on the teacher's dominant modes of variation.

\noindent\textbf{Why feature-space eigenvectors.}
For linear kernels, the sample-space ($n \times n$) and feature-space ($d \times d$) views share their eigenvalues (\cref{eq:duality}), so $\gL_{\mathrm{mag}}$ is the same in both. They differ only in the eigenvectors that $\gL_{\mathrm{dir}}$ aligns. Existing DPP-based objectives for generative models, including GDPP~\cite{elfeki2019gdpp}, DivGen~\cite{harrington2026divgen} and OSCAR~\cite{oscar}, build the $n \times n$ kernel of a batch; to our knowledge, we are the first to match eigenvectors in feature space. 

Specifically, the relation between the two views is explicit. Substituting $\vv_m = \tilde{\mPhi}^\top \vu_m / \sqrt{\lambda_m}$ gives
\begin{align}
  \langle \vv^S_m, \vv^T_m \rangle
  &= \frac{(\vu^S_m)^\top \, \tilde{\mPhi}_S \tilde{\mPhi}_T^\top \, \vu^T_m}{\sqrt{\lambda^S_m \lambda^T_m}},
  \nonumber \\
  \langle \vu^S_m, \vu^T_m \rangle
  &= (\vu^S_m)^\top \, \mI \, \vu^T_m .
  \label{eq:crossgram}
\end{align}
The feature-space alignment compares the two sets through their cross-similarities $\tilde{\mPhi}_S \tilde{\mPhi}_T^\top$, i.e., through what the samples look like. The sample-space alignment replaces this matrix by the identity, which assumes that the $i$-th student sample plays the role of the $i$-th teacher sample.

% The following Proposition answers why the feature space eigenvectors are more useful.
% The proof is provided in \cref{app:permutation}.

% The identity in \cref{eq:crossgram} is the crux: comparing $\vu^S_m$ with $\vu^T_m$ treats the $i$-th student sample as the counterpart of the $i$-th teacher sample. This holds only if both sets were generated from the same noises; for any other reference set, such as samples from other noises, another model or real images, the order of the samples is arbitrary. A useful objective should therefore depend on the two sets, not on how their samples are indexed. \Cref{prop:perm} shows that the feature-space alignment has this property and that the sample-space alignment does not (proof in \cref{app:permutation}).

Comparing $\vu^S_m$ with $\vu^T_m$ treats the $i$-th student and teacher samples as counterparts, which holds only for sets generated from the same noises. Proposition~\ref{prop:perm} shows that the feature-space alignment avoids this dependence on sample order (proof in \cref{app:permutation}).

\begin{restatable}[]{prop}{perm}
\label{prop:perm}
Let $\mPi$ be an $n \times n$ permutation matrix. 
Feature-space spectral alignment is invariant to sample permutations, whereas sample-space spectral alignment is permutation-dependent. In particular, if $\mPi$ is uniformly random, then for any unit vectors $\va,\vb$ satisfying $\vone^\top\vb=0$,
%
% (i) The feature-space eigen-pairs of $\mPi\tilde{\mPhi}_S$ equal those of $\tilde{\mPhi}_S$, so $\gL_{\mathrm{mag}}$ and $\gL_{\mathrm{dir}}$ depend only on the set of student samples, not on their order. (ii) The sample-space eigenvectors of $\mPi\tilde{\mPhi}_S$ are $\mPi\vu^S_m$, so the sample-space alignment changes with the order. (iii) If the order carries no information, i.e., $\mPi$ is uniformly random, then for unit vectors $\va, \vb$ with $\vone^\top \vb = 0$,
\begin{align}
  \E_{\mPi}\big[\langle \mPi\va, \vb \rangle\big] &= 0, \quad \text{and}\nonumber \\
  \E_{\mPi}\big[\langle \mPi\va, \vb \rangle^2\big]
  &%= \frac{1}{n} + \frac{1 - (\vone^\top\va)^2}{n(n-1)} 
  \le \frac{1}{n-1}.
  \label{eq:permbound}
\end{align}
\end{restatable}
% %
% \noindent\textit{Proof sketch.} (i) $(\mPi\tilde{\mPhi})^\top(\mPi\tilde{\mPhi}) = \tilde{\mPhi}^\top\tilde{\mPhi}$. (ii) $(\mPi\tilde{\mPhi})(\mPi\tilde{\mPhi})^\top = \mPi(\tilde{\mPhi}\tilde{\mPhi}^\top)\mPi^\top$. (iii) Under a uniform permutation, $\E[(\mPi\va)_i] = \vone^\top\va / n$, $\E[(\mPi\va)_i^2] = 1/n$ and $\E[(\mPi\va)_i(\mPi\va)_j] = ((\vone^\top\va)^2 - 1)/(n(n-1))$ for $i \ne j$. Summing against $\vb$ with $\sum_i b_i = 0$ and $\sum_{i \ne j} b_i b_j = -1$ gives \cref{eq:permbound}.
%
%The two properties underlying this proposition are established in \cref{prop:feature_invariance,prop:sample_equivariance} and proof is provided in \cref{app:permutation}.

\begin{figure*}[!t]
  \centering
  \includegraphics[width=\textwidth]{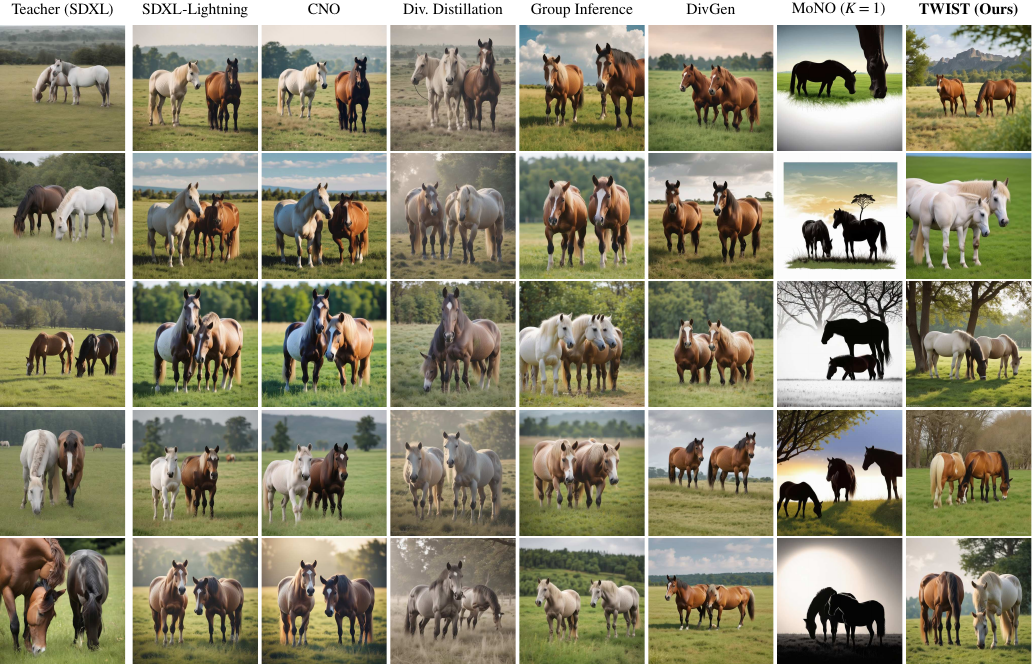}
  \caption{\textbf{Qualitative comparison on SDXL-Lightning} for the prompt ``a couple of horses grazing in a grass field''. TWIST uses selection from 224 candidates ($\alpha = 0.1$), the same cost as MoNO ($K{=}1$).}
  \label{fig:qual_horses}
\end{figure*}

Since the teacher set is centered at its mean, $\vone^\top\tilde{\mPhi}_T = \vzero$ and every teacher eigenvector with $\lambda^T_m > 0$ satisfies $\vone^\top\vu^T_m = 0$, so \cref{eq:permbound} applies with $\vb = \vu^T_m$. Without paired noises, the sample-space alignment is therefore noise of magnitude at most $1/\sqrt{n-1}$, whatever the actual structure of the two sets, and its gradient carries no information about directions of variation. The feature-space term has no such dependence. Empirically, a generator trained with the sample-space loss and independent noises does not move from the student, while the feature-space loss still increases diversity (\cref{app:additional}). We train with paired noises only because it lets us pre-compute each prompt's teacher set once.

\noindent
\textbf{Stable eigen-gradients.}
The gradient of an eigenvector involves $1/(\lambda_j - \lambda_i)$, which diverges for near-degenerate eigenvalues. These are common in the Gram matrix of a collapsed set. We use the exact eigen-decomposition in the forward pass and replace this factor by $(\lambda_j - \lambda_i) / ((\lambda_j - \lambda_i)^2 + \epsilon)$ in the backward pass, which is unchanged for well-separated eigenvalues and bounded otherwise.

\subsection{Spending Extra Budget on Forward Passes}
\label{sec:select}

Since TWIST needs no backward pass at inference, additional budget can be spent on forward passes alone. To compare with inference-time noise optimization at equal cost, we generate $N = Bn$ candidates per prompt in $B$ sets of $n$, each transformed by $\mathcal{W}_\psi$, and keep $n$ of them by greedy MAP inference (\cref{eq:greedy}) on the quality-weighted kernel
\begin{equation}
  \mL = \mathrm{diag}(\vq) \big(\mG\mG^\top + \epsilon \mI\big) \mathrm{diag}(\vq), \quad
  q_i = \exp(\alpha z_i),
  \label{eq:selkernel}
\end{equation}
where the rows of $\mG$ are $\ell_2$-normalized DINOv2 embeddings of the candidates, $z_i$ is the HPSv2~\cite{wu2023hps} score standardized over the candidates of the prompt, and $\alpha$ sets the quality-diversity trade-off. By \cref{sec:prelim_dpp}, MoNO optimizes the same greedy diversity term by gradient descent on the noise. Here it is evaluated over candidates already following the teacher's spectral shape, with an explicit quality weight.

\noindent
\textbf{Scoring with a tiny decoder.}
Decoding all $N$ candidates at full resolution dominates the cost of selection, yet both scorers only look at downsampled images. We therefore decode the candidates for scoring with the distilled tiny autoencoder TAESD-XL~\cite{bohan2023taesd}, and decode only the $n$ kept latents with the full VAE. The tiny decoder barely changes the scores and the selected sets (\cref{app:additional}). A candidate thus costs one pass of $\mathcal{W}_\psi$, one of $G_\eta$, one tiny decode and the two scorers, with no backpropagation.

\begin{figure}[t]
  \centering
  \includegraphics[width=\columnwidth]{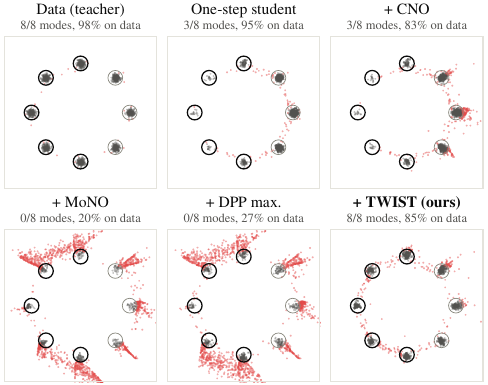}
  \caption{A one-step student trained on a tilted version of an 8-mode ring keeps 3 modes. We train a noise map with each objective and show samples at the end of training (red: samples off data modes; black rings: modes the student lost; results from first seed). Only the teacher-guided objective recovers all eight.}
  \label{fig:toy}
  \vspace{-0.3cm}
\end{figure}

\section{Experiments}
\label{sec:experiments}

\subsection{Setup}
\label{sec:setup}

\noindent
\textbf{Models and data.}
We apply TWIST to three distilled SDXL students: SDXL-Turbo~\cite{sauer2023add} (1 step, 512\,px), DMD2~\cite{yin2024dmd2} (1 step, 1024\,px) and SDXL-Lightning~\cite{lin2024lightning} (4 steps, 1024\,px). The teacher is SDXL-base~\cite{podell2024sdxl}. The generators are trained on teacher sets of prompts from a training prompt pool and evaluated on 128 held-out test prompts whose text does not appear in the training pool. For each test prompt and method we generate $n = 32$ images.

\noindent
\textbf{Metrics.}
Diversity and coverage are measured per prompt against 64 teacher samples of the same prompt, drawn from noises unused by any method, and averaged over prompts. We report recall and precision~\cite{kynkaanniemi2019improved} ($k = 3$) and the set's Vendi score~\cite{friedman2023vendi}, all on DINOv2-L features. Recall is the fraction of the teacher's samples falling within the student set's $k$-NN manifold: it measures how much of the teacher's per-prompt distribution the set covers. Quality and alignment are measured by PickScore~\cite{kirstain2023pick}, ImageReward~\cite{xu2023imagereward} and CLIPScore~\cite{hessel2021clipscore} (ViT-L/14). HPSv2 is the quality scorer inside our selection and is therefore not reported. To measure diversity and quality jointly, we also report the quality-weighted Vendi score (qVendi)~\cite{nguyen2024qvendi}, the Vendi score of the set multiplied by its mean quality, where the quality of an image is $\sigma(r)$ of its ImageReward score $r$. A set scores high only if it is both diverse and of high quality.

\noindent
\textbf{Baselines.}
We compare against inference-time noise optimization, CNO~\cite{kim2025cno}, DivGen~\cite{harrington2026divgen} and MoNO~\cite{shi2026mono} with $K = 1$ and $K = 4$ iterations, and against Group Inference~\cite{parmar2025gi} with 128 candidates, diversity distillation~\cite{gandikota2025distilling}, and an amortized variant of CNO in which our noise generator is trained with the CNO objective. All methods use the official settings or the settings reported for SDXL students; details are in the appendix.

\noindent
\textbf{Cost.}
We report the wall-clock time per kept image, from noise to pixels, including every decode, scoring and optimization step, measured on one NVIDIA B300 with a set of 32 images per call and the first prompt discarded as warm-up. For TWIST with selection, the pool size is chosen to stay at or below the time of MoNO with $K = 1$.

\subsection{Main Results}
\label{sec:main}

\begin{figure*}[t]
\centering
\begin{minipage}[t]{0.59\textwidth}
\vspace{0pt}
  \centering
  \setlength{\tabcolsep}{4.5pt}
  \resizebox{\linewidth}{!}{%
  \begin{tabular}{c l r ccccccc}
  \toprule
   & & & \multicolumn{4}{c}{Diversity / coverage} & \multicolumn{3}{c}{Quality / alignment} \\
\cmidrule(lr){4-7} \cmidrule(lr){8-10}
 & Method & s/img $\downarrow$ & Recall $\uparrow$ & Vendi $\uparrow$ & qVendi $\uparrow$ & Precision $\uparrow$ & PickScore $\uparrow$ & ImageReward $\uparrow$ & CLIP $\uparrow$ \\
\midrule
\multicolumn{2}{l}{Teacher (SDXL, 50 steps)} & -- & 0.826 & 5.62 & 3.49 & 0.800 & 22.28 & 0.662 & 26.32 \\
\midrule
 & Student & 0.062 & 0.114 & 3.30 & 2.12 & 0.577 & 22.39 & 0.718 & 26.28 \\
 & CNO & 0.109 & 0.132 & 3.44 & 2.20 & 0.569 & 22.38 & 0.713 & 26.24 \\
 & Amortized CNO & 0.063 & 0.232 & 4.53 & 2.81 & 0.426 & 22.02 & 0.621 & 26.16 \\
 & Diversity distillation & 0.076 & 0.540 & 5.53 & 3.28 & 0.500 & 22.17 & 0.512 & 26.06 \\
 & Group Inference & 0.275 & 0.157 & 3.89 & 2.48 & 0.539 & 22.33 & 0.699 & 26.53 \\
 & DivGen & 4.09 & 0.283 & 4.47 & 2.82 & 0.520 & 22.38 & 0.699 & 26.16 \\
 & MoNO ($K{=}1$) & 0.622 & 0.262 & 4.39 & 2.76 & 0.535 & 22.29 & 0.676 & 26.10 \\
 & MoNO ($K{=}4$) & 1.80 & 0.529 & 5.82 & 3.50 & 0.424 & 22.07 & 0.558 & 25.93 \\
\rowcolor{oursbg}
 & TWIST & & 0.164 & 3.74 & 2.38 & 0.554 & 22.33 & 0.690 & 26.14 \\
\rowcolor{oursbg}
\multirow{-10}{*}{\rotatebox[origin=c]{90}{SDXL-Turbo}}
 & TWIST + sel.\ (512) & 0.550 & 0.445 & 5.53 & 3.56 & 0.437 & 22.38 & 0.723 & 26.12 \\
\midrule
 & Student & 0.323 & 0.214 & 3.82 & 2.53 & 0.559 & 22.10 & 0.790 & 26.75 \\
 & CNO & 0.508 & 0.259 & 4.01 & 2.64 & 0.554 & 22.11 & 0.780 & 26.69 \\
 & Amortized CNO & 0.325 & 0.339 & 4.85 & 3.02 & 0.463 & 21.80 & 0.626 & 26.39 \\
 & Diversity distillation$^\dagger$ & 0.498 & 0.550 & 5.35 & 3.40 & 0.573 & 22.40 & 0.702 & 26.44 \\
 & Group Inference & 1.32 & 0.296 & 4.52 & 2.95 & 0.501 & 22.02 & 0.764 & 27.10 \\
 & DivGen & 15.5 & 0.327 & 4.51 & 2.94 & 0.550 & 22.14 & 0.777 & 26.59 \\
 & MoNO ($K{=}1$) & 2.01 & 0.455 & 5.26 & 3.28 & 0.467 & 21.85 & 0.648 & 26.52 \\
 & MoNO ($K{=}4$) & 5.84 & 0.647 & 6.71 & 3.93 & 0.367 & 21.57 & 0.468 & 26.11 \\
\rowcolor{oursbg}
 & TWIST & & 0.332 & 4.79 & 2.99 & 0.481 & 21.89 & 0.668 & 26.61 \\
\rowcolor{oursbg}
\multirow{-10}{*}{\rotatebox[origin=c]{90}{DMD2}}
 & TWIST + sel.\ (288) & 1.22 & 0.613 & 6.71 & 4.23 & 0.350 & 21.96 & 0.696 & 26.57 \\
\midrule
 & Student & 0.507 & 0.454 & 4.72 & 2.99 & 0.625 & 22.55 & 0.697 & 26.07 \\
 & CNO & 0.684 & 0.500 & 4.98 & 3.13 & 0.594 & 22.50 & 0.675 & 26.03 \\
 & Amortized CNO$^\ddagger$ & -- & -- & -- & -- & -- & -- & -- & -- \\
 & Diversity distillation & 0.498 & 0.688 & 6.61 & 3.64 & 0.437 & 21.94 & 0.307 & 25.89 \\
 & Group Inference & 2.05 & 0.625 & 5.96 & 3.68 & 0.537 & 22.44 & 0.640 & 26.35 \\
 & DivGen & 17.7 & 0.619 & 5.86 & 3.58 & 0.545 & 22.43 & 0.609 & 25.84 \\
 & MoNO ($K{=}1$) & 2.32 & 0.785 & 8.78 & 4.38 & 0.260 & 21.29 & 0.056 & 25.07 \\
 & MoNO ($K{=}4$) & 6.12 & 0.888 & 13.15 & 4.22 & 0.093 & 19.88 & $-$0.874 & 22.48 \\
\rowcolor{oursbg}
 & TWIST & & 0.538 & 5.13 & 3.19 & 0.599 & 22.46 & 0.648 & 26.01 \\
\rowcolor{oursbg}
\multirow{-10}{*}{\rotatebox[origin=c]{90}{SDXL-Lightning}}
 & TWIST + sel.\ (224) & 2.29 & 0.805 & 7.54 & 4.76 & 0.430 & 22.49 & 0.712 & 26.00 \\
  \bottomrule
  \end{tabular}%
  }
  \captionof{table}{Results on 128 test prompts, 32 images per prompt. ``TWIST + sel.\ ($N$)'' selects 32 of $N$ candidates with the quality-weighted DPP of \cref{sec:select}, scored on tiny-decoder images. qVendi: quality-weighted Vendi (\cref{sec:setup}). s/img: wall-clock seconds per kept image from noise to pixels. $^\dagger$Uses the 4-step DMD2 checkpoint after the teacher's first step. $^\ddagger$Training diverged and produced off-prompt images.}
  \label{tab:main}
\end{minipage}\hfill
\begin{minipage}[t]{0.39\textwidth}
\vspace{0pt}
  \centering
  \includegraphics[width=\linewidth]{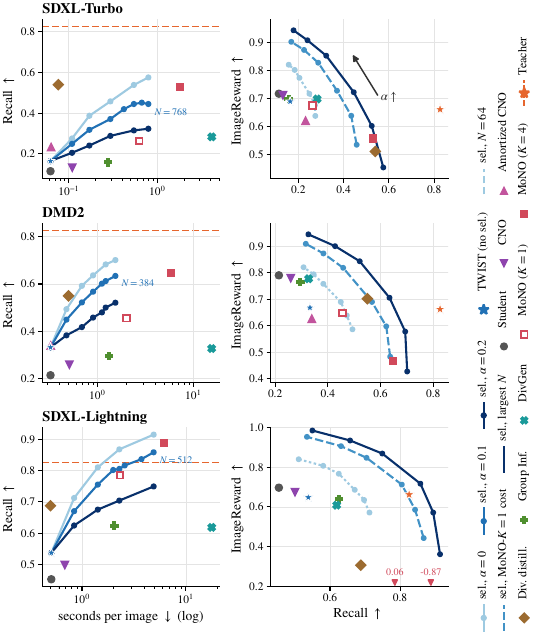}
  \caption{\textbf{Cost (left) and quality--diversity trade-off (right).}
  \emph{Left:} recall as the candidate pool grows from 32 (TWIST alone, $\star$) to $N$; dashed: teacher.
  \emph{Right:} each curve fixes the pool and raises $\alpha$ from 0 to 0.5.
  Baselines as in \Cref{tab:main}; MoNO's ImageReward on SDXL-Lightning is below the axis.}
  \label{fig:curves}
\end{minipage}
\end{figure*}

\Cref{tab:main} summarizes the results. Without selection, TWIST raises recall over every student at negligible cost: by 44\% on SDXL-Turbo, 55\% on DMD2 and 18\% on SDXL-Lightning. Amortized CNO, the other method without test-time optimization, trained with a repulsion objective on the same generator, loses considerably more quality on SDXL-Turbo and DMD2 and diverges on SDXL-Lightning.

At equal or lower cost than MoNO with $K = 1$, TWIST with selection covers more of the teacher's distribution on all three students (recall $0.445$ vs.\ $0.262$, $0.613$ vs.\ $0.455$ at 61\% of MoNO's time, and $0.805$ vs.\ $0.785$), with higher PickScore and ImageReward than MoNO. Its ImageReward stays above the student's on SDXL-Turbo and SDXL-Lightning, whereas MoNO lowers it on every model, from $0.70$ to $0.06$ on SDXL-Lightning. A larger quality weight (\cref{tab:ablation_sel}) gives the highest ImageReward of all methods on every student, with recall still above MoNO on SDXL-Turbo and DMD2. MoNO with $K = 4$ reaches a higher recall, but at $2.7$--$4.8\times$ the cost and with a large drop in quality; on SDXL-Lightning its precision falls to $0.09$, so its high Vendi score reflects artifacts. DivGen and Group Inference stay below TWIST in recall at a comparable or much larger cost. TWIST with selection also has the highest qVendi of all methods on every student ($3.56$, $4.23$ and $4.76$, vs.\ at most $3.50$, $3.93$ and $4.38$).

\Cref{fig:curves} extends the comparison to all budgets and quality weights. Recall grows steadily with the pool size (left); with $\alpha = 0$, TWIST with selection exceeds even MoNO with $K = 4$ in recall at a lower cost. Every baseline lies below the curve of the largest pool (right), and the curve at MoNO's cost already dominates MoNO with $K = 1$. Diversity distillation reaches a high recall at almost no cost, but at a lower ImageReward than TWIST at the same recall; on DMD2, where it uses the 4-step checkpoint, it is the strongest baseline. In \Cref{fig:qual_horses}, the baselines repeat two horses facing the camera and MoNO turns them into silhouettes, while TWIST varies the scene and pose, as the teacher does.

Our precision is lower than that of the students and of MoNO on SDXL-Turbo and DMD2. Precision rewards sets that stay near a few dense modes of the teacher's $k$-NN manifold, so it falls for every method as the set spreads.

\subsection{Recovering Lost Modes in 2D}
\label{sec:toy}

To isolate the objective, we use a 2D example (\cref{fig:toy}). The data is a ring of eight Gaussian modes with equal weights. A one-step student is distilled from a flow trained on a tilted version of the data and keeps only three modes. We freeze the student and train a noise map in front of it, as in TWIST, with each objective on random-Fourier features of the samples: our teacher-guided spectral loss, CNO, MoNO and DPP volume maximization. Over five seeds, CNO keeps the three modes of the student, while MoNO and DPP volume maximization push most samples off the data (19\% and 26\% of the samples within three standard deviations of a mode). Our objective recovers all eight modes with 84\% of the samples on the data. Since it is the only objective with a reference, it can tell which directions of spread are meaningful.

\subsection{Ablation study}
\label{sec:ablation}

All ablations use SDXL-Turbo and the 128 test prompts unless stated otherwise.

\begin{table}[t]
\centering
\small
\resizebox{\linewidth}{!}{
\begin{tabular}{l ccccc}
\toprule
 & Rec.\ $\uparrow$ & Vendi $\uparrow$ & Prec.\ $\uparrow$ & Pick $\uparrow$ & IR $\uparrow$ \\
\midrule
Student & 0.114 & 3.30 & 0.577 & 22.39 & 0.718 \\
\midrule
Eigenvalues only ($\gL_{\mathrm{mag}}$) & 0.235 & 4.50 & 0.463 & 22.08 & 0.563 \\
Eigenvectors only ($\gL_{\mathrm{dir}}$) & 0.143 & 3.56 & 0.546 & 22.34 & 0.698 \\
Full (ours) & 0.164 & 3.74 & 0.554 & 22.33 & 0.690 \\
\bottomrule
\end{tabular}
}
\caption{Objective ablation on SDXL-Turbo (generator only, no selection).}
\label{tab:ablation_obj}
\vspace{-0.7cm}
\end{table}

\noindent
\textbf{Objective.}
Matching the eigenvalues alone increases recall more than the full loss, but at a large cost in quality (\cref{tab:ablation_obj}): it widens the set without controlling the directions. Matching eigenvectors alone keeps the quality but barely increases diversity. The full loss lies between the two with a much smaller loss of quality. The comparison is clearer at matched recall: with 64-candidate selection, the full loss and the eigenvalue-only loss reach the same recall ($0.260$ vs.\ $0.259$), and the full loss has higher precision ($+0.047 \pm 0.010$) and PickScore ($+0.054 \pm 0.018$).

\noindent
\textbf{Feature space.}
Computing the loss on DINOv2 CLS embeddings or on SDXL latents gives slightly higher recall but loses much more quality: the global embedding and the latent carry little information about which variations preserve the content of the image, whereas patch tokens describe the spatial layout.

\noindent
\textbf{Further ablations.}
The appendix (\cref{app:additional}) reports three further experiments. Without paired noises, the sample-space loss leaves the generator at the student, as predicted by Proposition~\ref{prop:perm}, while the feature-space loss still increases recall. Images can be generated one at a time, with the set mean estimated from a fixed bank of auxiliary noises, with no change in the results. Scoring the candidates on tiny-decoder images cuts the cost of selection by a third to a half with almost no change in the results.

\section{Conclusion}
\label{sec:conclusion}

We presented TWIST, a teacher-guided noise transform that restores the diversity of distilled diffusion models without changing their weights. Instead of pushing samples apart, it matches the spectrum of the student's set to the teacher's: the eigenvalues set how much the samples spread, and the eigenvectors set the directions, so the added diversity follows quality-preserving variations without a reward model or anchor. In 2D, this recovers every lost mode, and our ablations show that the eigenvectors are what preserve quality. TWIST costs one small U-Net pass at inference; with quality-weighted DPP selection, it reaches a higher recall than inference-time noise optimization at equal or lower cost on all three students, with higher PickScore and ImageReward than MoNO.

\section*{LLM Usage Statement}
Claude Code (Anthropic) was partially used for implementing several experiments, debugging the base code, computing evaluation metrics, and improving the figures' visuals. The text and figures' arrangements were refined with the help of Claude Opus 5.5, particularly to avoid ambiguous wording, inconsistencies in notations (in methodology), and flawed arrangement of text and table sections. Claude Opus 5.5 was also used to suggest condensed wording (rewritten by the authors) to meet the page limit. All ideas, experimental design, results and claims were produced and verified by the authors, who take full responsibility for the content.

% TWIST reaches a higher recall than inference-time noise optimization at equal or lower cost on SDXL-Turbo, DMD2 and SDXL-Lightning, while keeping PickScore and ImageReward at or above those of the student.

% \input{sec/2_formatting}
% \input{sec/3_finalcopy}
{
    \small
    \bibliographystyle{ieeenat_fullname}
    \bibliography{main}
}

% WARNING: do not forget to delete the supplementary pages from your submission 
% \input{sec/X_suppl}

\clearpage
\appendix
\renewcommand{\dbltopfraction}{0.9}       % a table* may fill up to 90% of a page top
\setcounter{dbltopnumber}{3}              % up to 3 table* per page top
\renewcommand{\dblfloatpagefraction}{0.75}% float-only pages must be >=75% full
\renewcommand{\textfraction}{0.05}
% ============================================================== Supp. A: DPP / GDPP background
% Notation follows the main paper (\gY, \mL, \mPhi, \vf_i, ...). Labels are prefixed app: to avoid clashes with the
% main body; replace \cref{eq:twistloss} / \cref{sec:objective} / \cref{sec:select} below if those labels changed.
\section{Determinantal point processes}
\label{app:dpp}

\Cref{sec:prelim_dpp} introduces the DPP kernel, its volume, the spectral score of \cref{eq:dppscore}, the Gram--covariance duality of \cref{eq:duality} and GDPP. This section adds the parts of DPP theory that the main paper uses without stating them: the normalized distribution and why it repels (\cref{app:dpp_lens}), its spectral decomposition and the relation to the Vendi score (\cref{app:dpp_spec}), and the greedy MAP inference behind our selection (\cref{app:dpp_map}). \Cref{app:permutation} proves Proposition~\ref{prop:perm}.

\subsection{L-ensembles and repulsion}
\label{app:dpp_lens}

A DPP~\cite{kulesza2012dpp} over a ground set $\gY = \{1, \dots, N\}$ with positive semi-definite kernel $\mL \in \sR^{N \times N}$ (an L-ensemble) is
\begin{equation}
  P_{\mL}(\rmY = \gS) = \frac{\det(\mL_{\gS})}{\det(\mL + \mI)}, \qquad \gS \subseteq \gY,
  \label{eq:app_lensemble}
\end{equation}
where $\mL_{\gS}$ is the principal submatrix indexed by $\gS$ and $\det(\mL_{\emptyset}) = 1$. The normalizer follows from $\sum_{\gS \subseteq \gY} \det(\mL_{\gS}) = \det(\mL + \mI)$, which is why $\log\det(\mL + \mI)$ in \cref{eq:dppscore} is the log of the total squared volume of all subsets. The marginal kernel $\mK = \mL(\mL + \mI)^{-1}$ gives the inclusion probabilities $P(\gS \subseteq \rmY) = \det(\mK_{\gS})$; in particular,
\begin{equation}
  P(i, j \in \rmY) = \emK_{ii}\emK_{jj} - \emK_{ij}^2 \;\le\; P(i \in \rmY)\, P(j \in \rmY),
  \label{eq:app_repulsion}
\end{equation}
with equality only when $\emK_{ij} = 0$. Two items are less likely to appear together than independently, and the more similar they are, the stronger this effect: a DPP is a model of repulsion.

\subsection{Spectral decomposition and the Vendi score}
\label{app:dpp_spec}

With $\mL = \sum_{m=1}^{N} \lambda_m \vu_m \vu_m^\top$, a DPP is a mixture of elementary DPPs~\cite{kulesza2012dpp}:
\begin{align}
  P_{\mL} &= \frac{1}{\det(\mL + \mI)} \sum_{J \subseteq \gY} \Big(\prod_{m \in J} \lambda_m\Big) P^{V_J},
  \label{eq:app_mixture} \\
  \E\big[|\rmY|\big] &= \sum_{m=1}^{N} \frac{\lambda_m}{\lambda_m + 1} = \Tr(\mK),
  \label{eq:app_size}
\end{align}
where $P^{V_J}$ is the elementary DPP with marginal kernel $\sum_{m \in J} \vu_m \vu_m^\top$, which always draws exactly $|J|$ items. Sampling first keeps each eigenvector $\vu_m$ independently with probability $\lambda_m / (\lambda_m + 1)$ and then draws items from the span of the kept eigenvectors. The eigenvalues thus set how strongly each direction is represented and the eigenvectors which directions are available, the two roles that $\gL_{\mathrm{mag}}$ and $\gL_{\mathrm{dir}}$ of \cref{eq:twistloss} match separately (the latter through the feature-space eigenvectors of \cref{eq:duality}).

The Vendi score~\cite{friedman2023vendi} that we report is a function of the same spectrum. For unit features, $\Tr(\mL) = N$, and
\begin{equation}
  \mathrm{Vendi} = \exp\Big(-\sum_{m} \bar{\lambda}_m \log \bar{\lambda}_m\Big), \qquad \bar{\lambda}_m = \lambda_m / N,
  \label{eq:app_vendi}
\end{equation}
the exponential of the entropy of the normalized eigenvalues, \ie the effective number of distinct directions the set occupies. Like \cref{eq:dppscore}, it depends on the eigenvalues alone and is therefore blind to rotations of the set (\cref{sec:prelim_dpp}): a set that spreads off the data scores as high as one that spreads along it. This is why we report Vendi together with recall and precision against the teacher's samples, which do depend on where the samples lie.

\subsection{Greedy MAP inference}
\label{app:dpp_map}

Finding the most likely subset of size $k$, $\argmax_{|\gS| = k} \det(\mL_{\gS})$, is NP-hard~\cite{ko1995exact}. Since $\log \det(\mL_{\gS})$ is submodular~\cite{kulesza2012dpp}, it is usually approximated greedily~\cite{chen2018fastgreedy}. By the Schur complement,
\begin{equation}
  \det(\mL_{\gS \cup \{j\}}) = \det(\mL_{\gS}) \big( \emL_{jj} - \mL_{j\gS} \mL_{\gS}^{-1} \mL_{\gS j} \big),
  \label{eq:app_schur}
\end{equation}
so the greedy step adds the item with the largest Schur complement. For the quality-weighted linear kernel $\emL_{ij} = q_i \vf_i^\top \vf_j q_j$ (\cref{eq:selkernel} with $\epsilon = 0$), the Schur complement equals $q_j^2 \|(\mI - \mP_{\gS})\vf_j\|_2^2$, which gives \cref{eq:greedy}: each step trades the quality of a candidate against its distance from the span of the items already selected. With $q_i = \exp(\alpha z_i)$ as in \cref{eq:selkernel}, $\log q_j^2 = 2\alpha z_j$. Maintaining an incremental Cholesky factor of $\mL_{\gS}$~\cite{chen2018fastgreedy}, selecting $n$ of $N$ candidates costs $O(N n d)$, negligible next to generating the candidates.

For unit features, $\|(\mI - \mP_{\gS})\vf_j\|_2^2 = 1 - \|\mP_{\gS}\vf_j\|_2^2$. MoNO~\cite{shi2026mono} minimizes $\|\mP_{\gS}\vf_j\|_2^2$ for each new sample by gradient descent on its noise, so it is a continuous form of the diversity term of the greedy step: it moves each new noise until its sample leaves the span of the previous ones, whereas greedy selection picks, among samples that already exist, the one furthest from that span. Both act on the volume, and hence on the eigenvalues only.

\subsection{Proof of Proposition~\ref{prop:perm}}
\label{app:permutation}

We first state a property of uniformly random permutations (\cref{lem:app_perm}), which gives the last claim of Proposition~\ref{prop:perm}.

\begin{lemma}[Moments of a randomly permuted vector]
\label{lem:app_perm}
Let $\mPi$ be a uniformly random $n \times n$ permutation matrix. For any unit vectors $\va, \vb$ satisfying $\vone^\top\vb = 0$,
\begin{align*}
  \E_{\mPi}\big[\langle \mPi\va, \vb \rangle\big] &= 0, \\
  \E_{\mPi}\big[\langle \mPi\va, \vb \rangle^2\big] &= \frac{1}{n} + \frac{1 - (\vone^\top\va)^2}{n(n-1)} \le \frac{1}{n-1},
\end{align*}
with equality when $\vone^\top\va = 0$.
\end{lemma}

\begin{proof}
Let $\vx = \mPi\va$, so that $x_i = a_{\pi(i)}$ for a uniformly random permutation $\pi$, and let $s = \vone^\top\va$. Each $\pi(i)$ is uniform over $\{1, \dots, n\}$, and each pair $(\pi(i), \pi(j))$, $i \ne j$, is uniform over the ordered pairs of distinct indices. Hence, using $\|\va\| = 1$,
\begin{align*}
  \E[x_i] &= \frac{s}{n}, \qquad \E[x_i^2] = \frac{1}{n}, \\
  \E[x_i x_j] &= \frac{\sum_{k \ne l} a_k a_l}{n(n-1)} = \frac{s^2 - 1}{n(n-1)} \quad (i \ne j).
\end{align*}
The first gives $\E[\langle \vx, \vb \rangle] = \frac{s}{n}\, \vone^\top\vb = 0$. For the second moment, $\|\vb\| = 1$ and $\vone^\top\vb = 0$ give $\sum_{i \ne j} b_i b_j = (\vone^\top\vb)^2 - \|\vb\|^2 = -1$, so
\begin{align*}
  \E\big[\langle \vx, \vb \rangle^2\big]
  &= \sum_i b_i^2 \, \E[x_i^2] + \sum_{i \ne j} b_i b_j \, \E[x_i x_j] \\
  &= \frac{1}{n} + \frac{1 - s^2}{n(n-1)}
  \le \frac{1}{n} + \frac{1}{n(n-1)} = \frac{1}{n-1},
\end{align*}
with equality when $s = 0$.
\end{proof}

Following \cref{lem:app_perm}, we now prove Proposition~\ref{prop:perm}.

\perm*

\begin{proof}
Let $\mPi$ permute the student samples, so that $\mPhi_S$ becomes $\mPi\mPhi_S$. Since $\mPi\vone = \vone$, centering at the teacher's mean commutes with the permutation: $\mPi\mPhi_S - \vone\vmu_T^\top = \mPi\tilde{\mPhi}_S$. We assume distinct nonzero eigenvalues, so that each eigenvector is defined up to its sign, which the absolute value in \cref{eq:twistloss} removes. We prove the three claims in order.

\emph{(i) Feature space.} The second-moment matrix is unchanged,
\begin{equation*}
  (\mPi\tilde{\mPhi}_S)^\top(\mPi\tilde{\mPhi}_S) = \tilde{\mPhi}_S^\top \mPi^\top\mPi \tilde{\mPhi}_S = \tilde{\mPhi}_S^\top\tilde{\mPhi}_S,
\end{equation*}
so the feature-space eigen-pairs $(\lambda^S_m, \vv^S_m)$, and with them $\gL_{\mathrm{mag}}$ and $\gL_{\mathrm{dir}}$, do not depend on the order of the samples. The same holds when the eigen-pairs are computed through the Gram matrix. Its eigenvectors become $\mPi\vu^S_m$ by (ii), and $\vv^S_m = (\mPi\tilde{\mPhi}_S)^\top \mPi\vu^S_m / \sqrt{\lambda^S_m} = \tilde{\mPhi}_S^\top\vu^S_m / \sqrt{\lambda^S_m}$ is unchanged.

\emph{(ii) Sample space.} The Gram matrix becomes
\begin{equation*}
  (\mPi\tilde{\mPhi}_S)(\mPi\tilde{\mPhi}_S)^\top = \mPi \, \tilde{\mPhi}_S\tilde{\mPhi}_S^\top \mPi^\top,
\end{equation*}
which has the same eigenvalues and the eigenvectors $\mPi\vu^S_m$. The sample-space alignment $\langle \mPi\vu^S_m, \vu^T_m \rangle$ therefore changes with $\mPi$ unless $\vu^T_m$ is itself invariant to the permutation.

\emph{(iii) Random order.} This is \cref{lem:app_perm}.
\end{proof}

\paragraph{Application in the main paper.}
The bound is applied with $\va = \vu^S_m$ and $\vb = \vu^T_m$. The teacher set is centered at its own mean, so $\vone^\top\tilde{\mPhi}_T = \vzero$, and every teacher eigenvector with $\lambda^T_m > 0$ satisfies $\vu^T_m = \tilde{\mPhi}_T\vv^T_m / \sqrt{\lambda^T_m}$ and hence $\vone^\top\vu^T_m = 0$. With unpaired noises the order of the student samples carries no information about the teacher's, so the sample-space alignment of each pair of eigenvectors has mean zero and standard deviation at most $1/\sqrt{n-1}$ ($0.18$ for $n = 32$), whatever the structure of the two sets.

\section{Method and training details}
\label{app:details}

\subsection{Noise generator}
\label{app:arch}

\paragraph{Architecture.}
The proposal network $\delta_\psi$ is a convolutional U-Net on the $4$-channel latent grid of the student's noise ($64{\times}64$ for SDXL-Turbo, $128{\times}128$ for DMD2 and SDXL-Lightning). An input $3{\times}3$ convolution maps the noise to $64$ channels, followed by three resolution levels with $64$, $128$ and $256$ channels. Each level has one residual block (GroupNorm, SiLU, $3{\times}3$ convolution, twice, with a $1{\times}1$ skip convolution when the width changes) and a stride-2 $3{\times}3$ convolution for downsampling. The bottleneck has two residual blocks at $256$ channels around a $4$-head self-attention layer over the spatial positions. The decoder mirrors the encoder with nearest-neighbor upsampling, a $3{\times}3$ convolution and a residual block on the concatenation with the skip connection. The output layer (GroupNorm, SiLU, $3{\times}3$ convolution to $4$ channels) is zero-initialized, so that $\delta_\psi \equiv 0$ and the student is unchanged at the start of training. The prompt enters through FiLM~\cite{perez2018film}: the pooled text embedding of SDXL's second text encoder ($\ve_{\vc} \in \sR^{1280}$) passes through a linear layer with SiLU, and every residual block maps it to a per-channel scale and shift applied after its first convolution, $\vh \mapsto \vh \odot (1 + \vs) + \vb$. The network has $13.8$M parameters, about $0.5\%$ of the SDXL UNet, and the same architecture is used for all three students.

\paragraph{Transform.}
The low band of \cref{sec:transform} keeps the frequencies with $\sqrt{2(\omega_x^2 + \omega_y^2)} \le \rho$, in cycles per latent pixel, with $\rho = 0.15$; since the mask is binary and symmetric, $\mP_L$ is an orthogonal projector. The soft cap of \cref{eq:rotate} is $\theta_{\max} = 30^\circ$ for SDXL-Turbo and $15^\circ$ for DMD2 and SDXL-Lightning.

\paragraph{Generating one image at a time.}
\label{app:single}
The set mean of \Cref{sec:transform} involves only the proposals of $\delta_\psi$, not the student. To generate a single image, we pass its noise through $\delta_\psi$ together with a fixed bank of $K = 31$ auxiliary noises, drawn once from a fixed seed and shared by all prompts, and subtract the mean over the $K + 1$ proposals; only the image's own noise is then denoised and decoded. This costs $K$ extra passes of the small network per image and none of the student, and it matches the set-of-32 default on both SDXL-Turbo and SDXL-Lightning (\cref{tab:inference}).

\subsection{Training data}
\label{app:data}

\paragraph{Teacher sets.}
The teacher is SDXL-base~\cite{podell2024sdxl} at $1024$\,px with the deterministic 50-step Euler sampler and classifier-free guidance $5$, in fp16 with the VAE in fp32. For every training prompt we generate $32$ teacher samples and store their initial noises, drawn from per-sample seeds. The training prompts are $512$ prompts drawn at random from a pool of $4{,}800$ (the drawn subset contains $427$ MS-COCO 2014 training captions and $85$ ImageNet class prompts of the form ``a photo of a \{class\}.''). The 128 test prompts are MS-COCO 2014 validation captions ($76$), ImageNet class prompts ($46$) and PartiPrompts ($6$); none of their texts appears in the training pool.

\paragraph{Noise pairing.}
For SDXL-Turbo, the teacher's $128{\times}128$ noise is average-pooled over $2{\times}2$ blocks and multiplied by $2$ (\cref{sec:objective}). The evaluation uses the same pairing for every method.

\subsection{Objective and optimization}
\label{app:train}

\paragraph{Features.}
The feature extractor is DINOv2-B~\cite{oquab2024dinov2} at $256$\,px, with the patch tokens taken after the final layer norm. The teacher's features are computed once and cached. During training, the latents are decoded with the tiny autoencoder TAESD-XL~\cite{bohan2023taesd}, which reduces the memory and time of the backward pass through the decoder; at evaluation, all images are decoded with the full SDXL VAE (candidates for selection are scored on TAESD-XL images, \cref{sec:select}).

\paragraph{Loss.}
We minimize \cref{eq:total} with $\beta = 10$ and $\gamma = 1$ (\cref{eq:kl}). The eigenvalues enter $\gL_{\mathrm{mag}}$ unnormalized, all $n = 32$ eigen-pairs of each set are used, and the stabilizer of the eigenvector gradient is $\epsilon = 10^{-10}$.

\paragraph{Gradient path.}
For the one-step students (SDXL-Turbo, DMD2), the gradient flows through the full generation. For the four-step SDXL-Lightning, we follow MoNO~\cite{shi2026mono} and use the student's one-step clean estimate at its first timestep as a proxy for the final sample during training, so each training sample needs one student pass instead of four. At inference, the trained transform is used with the student's standard four-step sampler.

\paragraph{Exact chunked gradient.}
The loss couples all $n$ samples of a set through the eigen-decomposition, but the backward pass through the student and the decoder only needs the gradient with respect to each sample's features. For the $1024$\,px students we therefore compute the features of the whole set without a graph, evaluate the loss and its gradient $\partial \gL / \partial \mPhi_S$, and then recompute the graph in chunks of $4$ samples and backpropagate $\partial \gL / \partial \mPhi_S$ into $\psi$. This gives the same gradient as a single pass at a fraction of the memory. The student uses gradient checkpointing throughout.

\paragraph{Optimization.}
\Cref{tab:hparams} lists all settings. We use Adam ($\beta_1 = 0.9$, $\beta_2 = 0.999$, no weight decay), a 20-step linear warm-up followed by a constant learning rate, and gradient-norm clipping at $1$. Each step uses $2$ prompts with their full sets of $32$ samples. The student runs in bf16 and the decoder in fp32. Only $\psi$ is trained; the student, the decoder and DINOv2 are frozen.

\begin{table}[t]
\centering
\footnotesize
\setlength{\tabcolsep}{3pt}
\resizebox{\columnwidth}{!}{%
\begin{tabular}{l ccc}
\toprule
 & Turbo & DMD2 & Lightning \\
\midrule
Steps / resolution & 1 / 512\,px & 1 / 1024\,px & 4 / 1024\,px \\
Noise latent & $4{\times}64^2$ & $4{\times}128^2$ & $4{\times}128^2$ \\
Rotation cap $\theta_{\max}$ & $30^\circ$ & $15^\circ$ & $15^\circ$ \\
Low-band cut-off $\rho$ & \multicolumn{3}{c}{0.15} \\
Centering & \multicolumn{3}{c}{set + spatial} \\
\midrule
$\beta$ / dir.\ weight / $\gamma$ & \multicolumn{3}{c}{10 / 1 / 1} \\
Features & \multicolumn{3}{c}{DINOv2-B patch tokens, 256\,px} \\
Decoder in training & \multicolumn{3}{c}{TAESD-XL} \\
Gradient path & full & full & one-step proxy \\
\midrule
Training prompts / $n$ & \multicolumn{3}{c}{512 / 32} \\
Learning rate & $2{\times}10^{-4}$ & $5{\times}10^{-5}$ & $5{\times}10^{-5}$ \\
Training steps & 400 & 400 & 500 \\
\bottomrule
\end{tabular}}
\caption{Training settings of the three noise generators (SDXL-Turbo, DMD2, SDXL-Lightning). All three use the same $13.8$M-parameter network, Adam with a 20-step warm-up and constant learning rate, gradient clipping at $1$ and bf16 for the student.}
\label{tab:hparams}
\end{table}

\paragraph{Checkpoint selection.}
Each generator is trained for $600$ steps and saved every $100$ steps. Every saved checkpoint is evaluated on a validation split of $48$ prompts, drawn at random from a held-out prompt pool and disjoint from both the training prompts and the $128$ test prompts, with $32$ samples per prompt for SDXL-Turbo and $16$ for DMD2 and SDXL-Lightning, and recall and precision computed against the teacher's samples as in the main evaluation. We keep the checkpoint with the highest validation recall that does not lose too much precision relative to the student, which is step $400$ for SDXL-Turbo and DMD2 and step $500$ for SDXL-Lightning (\cref{tab:hparams}). The test prompts are used only for the final evaluation.

\section{Ablations and additional results}
\label{app:additional}

Unless stated otherwise, all experiments in this section use SDXL-Turbo, the $128$ test prompts and $n = 32$ images per prompt, and are evaluated exactly as in \cref{tab:main}. Each ablation changes one component of the configuration of \cref{tab:hparams}; differences in the training setup of individual runs are stated in the table captions.

% ------------------------------------------------------------------------------------------------ objective
\subsection{Objective}
\label{app:abl_obj}

\begin{table}[t]
\centering
\footnotesize
\setlength{\tabcolsep}{3pt}
\resizebox{\columnwidth}{!}{%
\begin{tabular}{lccccc}
\toprule
 & Rec.\ $\uparrow$ & Vendi $\uparrow$ & Prec.\ $\uparrow$ & Pick $\uparrow$ & IR $\uparrow$ \\
\midrule
Student & 0.114 & 3.30 & 0.577 & 22.39 & 0.718 \\
\midrule
\multicolumn{6}{l}{\textit{Generator only}} \\
Eigenvalues only ($\gL_{\mathrm{mag}}$) & 0.235 & 4.50 & 0.463 & 22.08 & 0.563 \\
Eigenvectors only ($\gL_{\mathrm{dir}}$) & 0.143 & 3.56 & 0.546 & 22.34 & 0.698 \\
Full (\cref{tab:main} model) & 0.164 & 3.74 & 0.554 & 22.33 & 0.690 \\
\midrule
\multicolumn{6}{l}{\textit{Selection of 32 from 64 candidates}} \\
Student, $\alpha = 0.1$ & 0.157 & 3.76 & 0.547 & 22.44 & 0.754 \\
Eigenvalues only, $\alpha = 0.1$ & 0.343 & 5.32 & 0.414 & 22.10 & 0.580 \\
Eigenvalues only, $\alpha = 0.3$ & 0.259 & 4.58 & 0.468 & 22.30 & 0.707 \\
Eigenvectors only, $\alpha = 0.1$ & 0.218 & 4.12 & 0.520 & 22.37 & 0.728 \\
Full (\cref{tab:main} model), $\alpha = 0.1$ & 0.260 & 4.39 & 0.515 & 22.35 & 0.714 \\
\bottomrule
\end{tabular}}
\caption{\textbf{Objective ablation} on SDXL-Turbo. Selection uses the quality-weighted DPP of \cref{sec:select} on full-VAE images.}
\label{tab:supp_obj}
\end{table}

\Cref{tab:supp_obj} extends the objective ablation of the main paper (\cref{tab:ablation_obj}). The full loss lies between the two single terms in recall, Vendi and ImageReward, and has the highest precision of the three: matching the eigenvalues alone gives the largest increase in recall and Vendi but the largest loss in precision and ImageReward, and matching the eigenvectors alone keeps most of the quality of the student but moves its diversity only slightly.

The second block compares the losses at matched recall. With $64$ candidates, the eigenvalue-only generator reaches the recall of the full loss only with a larger quality weight ($\alpha = 0.3$ vs.\ $0.1$, recall $0.259$ vs.\ $0.260$); at that point, the full loss (the model of \cref{tab:main}) has a higher precision ($+0.047$), PickScore ($+0.05$), ImageReward ($+0.007$) for a slightly lower Vendi. At the same $\alpha = 0.1$, the eigenvalue-only pool has the higher recall but loses $0.13$ in ImageReward. The eigenvector-only pool stays below the full loss in recall at the same $\alpha$.

% ------------------------------------------------------------------------------------------------ pairing
\subsection{Sample-space vs.\ feature-space alignment and noise pairing}
\label{app:abl_pair}

\begin{table}[t]
\centering
\footnotesize
\setlength{\tabcolsep}{3pt}
\begin{tabular}{l ccccc}
\toprule
 & Rec.\ $\uparrow$ & Vendi $\uparrow$ & Prec.\ $\uparrow$ & Pick $\uparrow$ & IR $\uparrow$ \\
\midrule
Student & 0.114 & 3.30 & 0.577 & 22.39 & 0.718 \\
\midrule
\multicolumn{6}{l}{\textit{Paired noises}} \\
$n{\times}n$ & 0.175 & 3.87 & 0.537 & 22.30 & 0.678 \\
$d{\times}d$ (ours) & 0.164 & 3.74 & 0.554 & 22.33 & 0.690 \\
\midrule
\multicolumn{6}{l}{\textit{Unpaired noises}} \\
$n{\times}n$ & 0.121 & 3.32 & 0.577 & 22.39 & 0.721 \\
$d{\times}d$ & 0.185 & 3.89 & 0.517 & 22.23 & 0.679 \\
\bottomrule
\end{tabular}
\caption{\textbf{Alignment view and noise pairing} on SDXL-Turbo. $n{\times}n$: eigenvectors of the sample-space Gram matrices; $d{\times}d$: feature-space eigenvectors (\cref{sec:objective}). Paired: the student receives the noises of the teacher's samples. Unpaired: the student receives fresh noises, independent of the teacher's; the unpaired runs were trained for $300$ steps.}
\label{tab:supp_pair}
\end{table}

\Cref{tab:supp_pair} tests Proposition~\ref{prop:perm}. Without paired noises, the order of the student's samples carries no information about the order of the teacher's, and the sample-space alignment is noise of magnitude at most $1/\sqrt{n-1}$ (\cref{app:permutation}). The unpaired $n{\times}n$ generator accordingly does not leave the student: recall, precision and ImageReward are within $0.01$ of the student's and Vendi within $0.02$. The unpaired $d{\times}d$ generator, whose loss does not depend on the order of the samples, does move away from the student, with a recall of $0.185$ against the student's $0.114$. With paired noises, where the order of the samples is meaningful, the two views behave alike. Only the order-invariant feature-space loss learns from unpaired noises, which makes it applicable to reference sets that are not generated from the student's noises.

% ------------------------------------------------------------------------------------------------ feature space
\subsection{Feature space}
\label{app:abl_space}

\begin{table}[t]
\centering
\footnotesize
\setlength{\tabcolsep}{3pt}
\resizebox{\columnwidth}{!}{%
\begin{tabular}{l c ccccc}
\toprule
Feature space & Sel. & Rec.\ $\uparrow$ & Vendi $\uparrow$ & Prec.\ $\uparrow$ & Pick $\uparrow$ & IR $\uparrow$ \\
\midrule
Student & -- & 0.114 & 3.30 & 0.577 & 22.39 & 0.718 \\
Student & 64 & 0.157 & 3.76 & 0.547 & 22.44 & 0.754 \\
\midrule
SDXL latent ($16{\times}16$) & -- & 0.184 & 4.22 & 0.449 & 21.88 & 0.553 \\
 & 64 & 0.298 & 4.94 & 0.403 & 21.94 & 0.579 \\
DINOv2-L CLS & -- & 0.187 & 4.22 & 0.458 & 22.00 & 0.560 \\
 & 64 & 0.300 & 4.93 & 0.423 & 22.03 & 0.577 \\
DINOv2-B patches (ours) & -- & 0.164 & 3.74 & 0.554 & 22.33 & 0.690 \\
 & 64 & 0.260 & 4.39 & 0.515 & 22.35 & 0.714 \\
\bottomrule
\end{tabular}}
\caption{\textbf{Feature space of the objective} on SDXL-Turbo, without selection and with selection of $32$ from $64$ candidates ($\alpha = 0.1$). All runs use the configuration of \cref{tab:main} and differ only in $\phi$.}
\label{tab:supp_space}
\end{table}

\Cref{tab:supp_space} compares the feature space in which the two sets are matched. The SDXL latent, average-pooled to $16{\times}16$, and the global CLS embedding of DINOv2-L give a slightly higher recall than the patch tokens of DINOv2-B ($0.19$ vs.\ $0.16$) but lose much more quality: precision falls by $0.12$ from the student instead of $0.02$, and ImageReward by $0.16$ instead of $0.03$. With selection, all feature spaces gain about $0.1$ in recall, and the ordering stays the same; only the patch-token generator keeps the ImageReward of the selected set close to the student's ($0.714$ vs.\ $0.718$, against at most $0.58$ for the other feature spaces).

The CLS ablation is, if anything, favored by the evaluation: it is trained on the DINOv2-L CLS embedding, the same feature space in which recall and precision are measured. That it still loses more quality than the patch tokens supports our explanation in the main paper: the global embedding and the latent carry little information about which variations preserve the content of the image, whereas patch tokens describe its spatial layout.

% ------------------------------------------------------------------------------------------------ inference variants
\subsection{Inference variants}
\label{app:abl_inference}

\begin{table}[t]
\centering
\footnotesize
\setlength{\tabcolsep}{3pt}
\resizebox{\columnwidth}{!}{%
\begin{tabular}{l r ccccc}
\toprule
 & s/img & Rec.\ $\uparrow$ & Vendi $\uparrow$ & Prec.\ $\uparrow$ & Pick $\uparrow$ & IR $\uparrow$ \\
\midrule
\multicolumn{7}{l}{\textit{Set mean, SDXL-Turbo}} \\
Set of 32 & & 0.164 & 3.74 & 0.554 & 22.33 & 0.690 \\
One at a time, bank & & 0.160 & 3.69 & 0.555 & 22.34 & 0.690 \\
One at a time, no mean & & 0.087 & 3.40 & 0.410 & 22.04 & 0.579 \\
\multicolumn{7}{l}{\textit{Set mean, SDXL-Lightning}} \\
Set of 32 & & 0.538 & 5.13 & 0.599 & 22.46 & 0.648 \\
One at a time, bank & & 0.538 & 5.11 & 0.597 & 22.46 & 0.655 \\
One at a time, no mean & & 0.424 & 6.16 & 0.301 & 21.35 & 0.121 \\
\midrule
\multicolumn{7}{l}{\textit{Decoder for scoring}} \\
Turbo, 256, VAE & 0.568 & 0.385 & 5.28 & 0.450 & 22.38 & 0.715 \\
Turbo, 256, tiny & 0.296 & 0.371 & 5.15 & 0.462 & 22.38 & 0.729 \\
Lightning, 128, VAE & 2.07 & 0.760 & 6.99 & 0.472 & 22.50 & 0.692 \\
Lightning, 128, tiny & 1.42 & 0.755 & 7.00 & 0.468 & 22.49 & 0.695 \\
\bottomrule
\end{tabular}}
\caption{\textbf{Inference variants.} Top: generating images one at a time, with the set mean of \cref{sec:transform} estimated from a fixed bank of $31$ auxiliary noises passed through $\delta_\psi$ only (\cref{app:single}), or dropped. Bottom: scoring the candidates on tiny-decoder images instead of full-VAE images, at the same pool size.}
\label{tab:inference}
\end{table}

\paragraph{Generating one image at a time.}
Passing an image's noise through $\delta_\psi$ together with the fixed bank of \Cref{app:single}, and denoising only the image's own noise, reproduces the results of generating the set jointly (\cref{tab:inference}). Dropping the set mean instead lets through a component shared by all noises, which the generator never had to suppress during training; it degrades quality and, on SDXL-Turbo, even reduces diversity.

\paragraph{Tiny decoder for scoring.}
Scoring the candidates on tiny-decoder images halves the cost of selection on SDXL-Turbo and cuts it by a third on SDXL-Lightning, with almost no change in the results at the same pool size (\cref{tab:inference}). Over the $128$ prompts of SDXL-Lightning, the DINOv2-B features of the tiny decoder's images have a mean cosine similarity of $0.993$ with those of the full decoder's, their HPSv2 scores a mean Spearman correlation of $0.98$, and $86$--$88\%$ of the selected images are the same.

% ------------------------------------------------------------------------------------------------ plain DPP

\begin{table*}[t]
\centering
\caption{\textbf{Plain DPP selection vs.\ selection from TWIST's candidates} (128 test prompts, $n{=}32$ kept samples per prompt).
Both select $n$ of $N$ candidates with the same greedy quality-weighted DPP ($q_i{=}\exp(\alpha z_i)$, $\alpha{=}0.1$, $z$ the standardized HPSv2 score; DINOv2-B kernel) and differ only in the candidate pool: the student's own samples (\emph{Student + DPP}) or TWIST's (\emph{TWIST + DPP}). $N/n{=}1\times$ is plain sampling without selection. All candidates are decoded with the full VAE.}
\label{tab:supp_dpp}
\small
\setlength{\tabcolsep}{4pt}

\begin{tabular}{ll c ccc cccc}
\toprule
& & & \multicolumn{3}{c}{Diversity} & \multicolumn{4}{c}{Quality / alignment} \\
\cmidrule(lr){4-6}\cmidrule(lr){7-10}
Model & Pool & $N/n$ & Rec.$\uparrow$ & Vendi$\uparrow$ & qVendi$\uparrow$ & Prec.$\uparrow$ & Pick$\uparrow$ & IR$\uparrow$ & CLIP$\uparrow$ \\
\midrule
 & Student & 1$\times$ & 0.114 & 3.30 & 2.12 & 0.577 & 22.39 & 0.718 & 26.28 \\
 & \cellcolor{oursbg}TWIST & \cellcolor{oursbg}1$\times$ & \cellcolor{oursbg}0.164 & \cellcolor{oursbg}3.74 & \cellcolor{oursbg}2.38 & \cellcolor{oursbg}0.554 & \cellcolor{oursbg}22.33 & \cellcolor{oursbg}0.690 & \cellcolor{oursbg}26.14 \\
\cmidrule(lr){2-10}
 & Student + DPP & 2$\times$ & 0.157 & 3.76 & 2.45 & 0.547 & 22.44 & 0.754 & 26.30 \\
 & \cellcolor{oursbg}TWIST + DPP & \cellcolor{oursbg}2$\times$ & \cellcolor{oursbg}0.260 & \cellcolor{oursbg}4.39 & \cellcolor{oursbg}2.81 & \cellcolor{oursbg}0.515 & \cellcolor{oursbg}22.35 & \cellcolor{oursbg}0.714 & \cellcolor{oursbg}26.17 \\
\cmidrule(lr){2-10}
 & Student + DPP & 4$\times$ & 0.220 & 4.12 & 2.68 & 0.520 & 22.46 & 0.755 & 26.32 \\
\multirow{-6}{*}{\rotatebox{90}{\scriptsize SDXL-Turbo}} & \cellcolor{oursbg}TWIST + DPP & \cellcolor{oursbg}4$\times$ & \cellcolor{oursbg}0.325 & \cellcolor{oursbg}4.83 & \cellcolor{oursbg}3.11 & \cellcolor{oursbg}0.480 & \cellcolor{oursbg}22.38 & \cellcolor{oursbg}0.724 & \cellcolor{oursbg}26.18 \\
\midrule
 & Student & 1$\times$ & 0.214 & 3.82 & 2.53 & 0.559 & 22.10 & 0.790 & 26.75 \\
 & \cellcolor{oursbg}TWIST & \cellcolor{oursbg}1$\times$ & \cellcolor{oursbg}0.332 & \cellcolor{oursbg}4.79 & \cellcolor{oursbg}2.99 & \cellcolor{oursbg}0.481 & \cellcolor{oursbg}21.89 & \cellcolor{oursbg}0.668 & \cellcolor{oursbg}26.61 \\
\cmidrule(lr){2-10}
 & Student + DPP & 2$\times$ & 0.282 & 4.32 & 2.89 & 0.511 & 22.15 & 0.830 & 26.80 \\
 & \cellcolor{oursbg}TWIST + DPP & \cellcolor{oursbg}2$\times$ & \cellcolor{oursbg}0.455 & \cellcolor{oursbg}5.61 & \cellcolor{oursbg}3.50 & \cellcolor{oursbg}0.418 & \cellcolor{oursbg}21.93 & \cellcolor{oursbg}0.684 & \cellcolor{oursbg}26.63 \\
\cmidrule(lr){2-10}
 & Student + DPP & 4$\times$ & 0.348 & 4.69 & 3.16 & 0.479 & 22.18 & 0.851 & 26.80 \\
\multirow{-6}{*}{\rotatebox{90}{\scriptsize DMD2}} & \cellcolor{oursbg}TWIST + DPP & \cellcolor{oursbg}4$\times$ & \cellcolor{oursbg}0.525 & \cellcolor{oursbg}6.13 & \cellcolor{oursbg}3.83 & \cellcolor{oursbg}0.373 & \cellcolor{oursbg}21.95 & \cellcolor{oursbg}0.692 & \cellcolor{oursbg}26.59 \\
\midrule
 & Student & 1$\times$ & 0.454 & 4.72 & 2.99 & 0.625 & 22.55 & 0.697 & 26.07 \\
 & \cellcolor{oursbg}TWIST & \cellcolor{oursbg}1$\times$ & \cellcolor{oursbg}0.538 & \cellcolor{oursbg}5.13 & \cellcolor{oursbg}3.19 & \cellcolor{oursbg}0.599 & \cellcolor{oursbg}22.46 & \cellcolor{oursbg}0.648 & \cellcolor{oursbg}26.01 \\
\cmidrule(lr){2-10}
 & Student + DPP & 2$\times$ & 0.594 & 5.63 & 3.61 & 0.550 & 22.59 & 0.745 & 26.08 \\
 & \cellcolor{oursbg}TWIST + DPP & \cellcolor{oursbg}2$\times$ & \cellcolor{oursbg}0.678 & \cellcolor{oursbg}6.20 & \cellcolor{oursbg}3.89 & \cellcolor{oursbg}0.520 & \cellcolor{oursbg}22.49 & \cellcolor{oursbg}0.683 & \cellcolor{oursbg}26.02 \\
\cmidrule(lr){2-10}
 & Student + DPP & 4$\times$ & 0.689 & 6.34 & 4.07 & 0.498 & 22.60 & 0.747 & 26.08 \\
\multirow{-6}{*}{\rotatebox{90}{\scriptsize SDXL-Lightning}} & \cellcolor{oursbg}TWIST + DPP & \cellcolor{oursbg}4$\times$ & \cellcolor{oursbg}0.760 & \cellcolor{oursbg}6.99 & \cellcolor{oursbg}4.39 & \cellcolor{oursbg}0.472 & \cellcolor{oursbg}22.50 & \cellcolor{oursbg}0.692 & \cellcolor{oursbg}25.98 \\
\bottomrule
\end{tabular}
\end{table*}  

\subsection{Is the gain due to selection alone?}
\label{app:abl_dpp}

TWIST with selection combines two components: the trained transform, which changes the candidates, and the quality-weighted DPP, which picks among them. \Cref{tab:supp_dpp} separates the two by applying the same selection to the student's own samples. At equal pool size, selecting from TWIST's candidates gives a higher recall, Vendi and qVendi than selecting from the student's on all three students, at both pool sizes. Selection alone does not undo the collapse: on SDXL-Turbo and DMD2, the student's pool at $4\times$ the candidates stays below TWIST's pool at $2\times$ in recall ($0.220$ vs.\ $0.260$ and $0.348$ vs.\ $0.455$) and in qVendi ($2.68$ vs.\ $2.81$ and $3.16$ vs.\ $3.50$). On SDXL-Lightning, whose student is the least collapsed, the student's pool at $4\times$ is slightly ahead of TWIST's at $2\times$, and TWIST's pool is ahead at equal size.

Selection from the student's pool keeps a higher ImageReward and PickScore, since it picks from samples of higher quality that differ little from each other; the quality weight $\alpha$ can move TWIST's pool along the same trade-off (\cref{fig:curves}). \Cref{fig:curves} (left) also reports how recall grows with the pool size for TWIST with selection.

% ------------------------------------------------------------------------------------------------ selection budget

\begin{table*}[t]
\centering
\caption{\textbf{Selection budget and quality weight} (128 test prompts, $n{=}32$ kept samples per prompt, same setting as \cref{tab:main}). ``TWIST + sel.\ ($N$)'' selects $n$ of $N$ candidates with the quality-weighted DPP of \Cref{sec:select} ($\alpha{=}0.1$); ``quality'' uses $\alpha{=}0.2$. s/img: median wall-clock seconds per image on one NVIDIA B300.}
\label{tab:ablation_sel}
\small
\setlength{\tabcolsep}{4pt}

\begin{tabular}{ll c ccc cccc}
\toprule
& & & \multicolumn{3}{c}{Diversity} & \multicolumn{4}{c}{Quality / alignment} \\
\cmidrule(lr){4-6}\cmidrule(lr){7-10}
Model & Method & s/img$\downarrow$ & Rec.$\uparrow$ & Vendi$\uparrow$ & qVendi$\uparrow$ & Prec.$\uparrow$ & Pick$\uparrow$ & IR$\uparrow$ & CLIP$\uparrow$ \\
\midrule
\multirow{4}{*}{\rotatebox{90}{\scriptsize SDXL-Turbo}}
 & TWIST & -- & 0.164 & 3.74 & 2.38 & 0.554 & 22.33 & 0.690 & 26.14 \\
 & TWIST + sel.\ (256) & 0.296 & 0.371 & 5.15 & 3.32 & 0.462 & 22.38 & 0.729 & 26.17 \\
 & TWIST + sel.\ (512) & 0.550 & 0.445 & 5.53 & 3.56 & 0.437 & 22.38 & 0.723 & 26.12 \\
 & TWIST + sel.\ (512, quality) & 0.550 & 0.315 & 4.70 & 3.15 & 0.496 & 22.59 & 0.847 & 26.25 \\
\midrule
\multirow{4}{*}{\rotatebox{90}{\scriptsize DMD2}}
 & TWIST & -- & 0.332 & 4.79 & 2.99 & 0.481 & 21.89 & 0.668 & 26.61 \\
 & TWIST + sel.\ (192) & 0.900 & 0.568 & 6.36 & 3.99 & 0.366 & 21.95 & 0.690 & 26.60 \\
 & TWIST + sel.\ (288) & 1.22 & 0.613 & 6.71 & 4.23 & 0.350 & 21.96 & 0.696 & 26.57 \\
 & TWIST + sel.\ (288, quality) & 1.22 & 0.500 & 5.66 & 3.75 & 0.430 & 22.20 & 0.830 & 26.73 \\
\midrule
\multirow{4}{*}{\rotatebox{90}{\scriptsize\shortstack{SDXL-\\Lightning}}}
 & TWIST & -- & 0.538 & 5.13 & 3.19 & 0.599 & 22.46 & 0.648 & 26.01 \\
 & TWIST + sel.\ (128) & 1.42 & 0.755 & 7.00 & 4.39 & 0.468 & 22.49 & 0.695 & 25.99 \\
 & TWIST + sel.\ (224) & 2.29 & 0.805 & 7.54 & 4.76 & 0.430 & 22.49 & 0.712 & 26.00 \\
 & TWIST + sel.\ (224, quality) & 2.29 & 0.704 & 6.52 & 4.32 & 0.499 & 22.69 & 0.847 & 26.15 \\
\bottomrule
\end{tabular}
\end{table*}

\subsection{Selection budget and quality weight}
\label{app:abl_sel}

\Cref{tab:ablation_sel} reports TWIST with selection at the two pool sizes per student used in \Cref{tab:main}, and with a larger quality weight. A larger pool raises recall, Vendi and qVendi with PickScore, ImageReward and CLIP score almost unchanged, while precision decreases (\eg $0.371 \to 0.445$ in recall from $256$ to $512$ candidates on SDXL-Turbo).

The quality weight $\alpha$ of \Cref{sec:select} moves the selection along the diversity--quality trade-off. Doubling it from $0.1$ to $0.2$ at the largest pool gives the highest ImageReward of all methods on every student ($0.847$, $0.830$ and $0.847$) and raises PickScore, CLIP score and precision, at a lower recall. That recall is still above MoNO with $K = 1$ on SDXL-Turbo ($0.315$ vs.\ $0.262$) and DMD2 ($0.500$ vs.\ $0.455$), and below it on SDXL-Lightning ($0.704$ vs.\ $0.785$).

% ------------------------------------------------------------------------------------------------ oracle

\begin{table*}[t]
\centering
\caption{\textbf{Inference-time oracle vs.\ MoNO at an equal budget} (SDXL-Turbo, 128 test prompts, $n{=}32$).
The oracle optimizes the TWIST objective (\cref{eq:total}) directly on the $32$ noises of each prompt, without $\delta_\psi$, against the teacher's samples \emph{from the same noises}. It uses information that is unavailable at inference and serves only as a reference for the objective. Like MoNO $K{=}1$, it takes one backward pass per sample: one Adam step on a free proposal that passes through the transform of \Cref{sec:transform} with the soft cap $\theta_{\max}$ (learning rate $0.1$ for $\theta_{\max}{=}30^\circ$ and $0.15$ for $45^\circ$). s/img: median wall-clock seconds per image on one NVIDIA B300. Bold: best among MoNO and the oracles.}
\label{tab:supp_oracle}
\small
\setlength{\tabcolsep}{4pt}

\begin{tabular}{l c ccc cccc}
\toprule
& & \multicolumn{3}{c}{Diversity} & \multicolumn{4}{c}{Quality / alignment} \\
\cmidrule(lr){3-5}\cmidrule(lr){6-9}
Method & s/img$\downarrow$ & Rec.$\uparrow$ & Vendi$\uparrow$ & qVendi$\uparrow$ & Prec.$\uparrow$ & Pick$\uparrow$ & IR$\uparrow$ & CLIP$\uparrow$ \\
\midrule
\textcolor{gray}{Teacher (reference)} & \textcolor{gray}{--} & \textcolor{gray}{0.826} & \textcolor{gray}{5.62} & \textcolor{gray}{3.49} & \textcolor{gray}{0.800} & \textcolor{gray}{22.28} & \textcolor{gray}{0.662} & \textcolor{gray}{26.32} \\
\midrule
Student & -- & 0.114 & 3.30 & 2.12 & 0.577 & 22.39 & 0.718 & 26.28 \\
TWIST & -- & 0.164 & 3.74 & 2.38 & 0.554 & 22.33 & 0.690 & 26.14 \\
\midrule
MoNO $K{=}1$ & 0.58 & 0.262 & \textbf{4.39} & 2.76 & 0.535 & 22.29 & 0.676 & 26.10 \\
Oracle, $\theta_{\max}{=}30^\circ$ & \textbf{0.32} & 0.221 & 4.06 & 2.61 & \textbf{0.560} & \textbf{22.37} & \textbf{0.738} & \textbf{26.15} \\
Oracle, $\theta_{\max}{=}45^\circ$ & \textbf{0.32} & \textbf{0.268} & 4.36 & \textbf{2.77} & 0.526 & 22.29 & 0.706 & 26.10 \\
\bottomrule
\end{tabular}
\end{table*}

\subsection{Inference-time oracle}
\label{app:abl_oracle}

Without selection, TWIST reaches a lower recall than MoNO with $K = 1$, which optimizes every noise at inference. To separate the effect of the objective from that of amortization, we optimize the TWIST objective directly on the noises of each test prompt at inference, against the teacher's samples from the same noises, with the same budget as MoNO with $K = 1$: one backward pass per sample. The update follows TWIST's own parameterization: a free proposal per noise, initialized at zero, passes through the transform of \cref{sec:transform} and takes one Adam step, and the cap $\theta_{\max}$ bounds the rotation of each noise. This oracle uses the teacher at inference and is not a practical method; it measures what the objective achieves when it does not have to generalize across prompts.

\Cref{tab:supp_oracle} shows that, at the same budget, the TWIST objective matches or improves on MoNO, with $\theta_{\max}$ setting where it lands on the diversity--quality trade-off. With $\theta_{\max} = 45^\circ$, the oracle reaches a very close Vendi score to MoNO ($4.36$ vs.\ $4.39$), a higher qVendi ($2.77$ vs.\ $2.76$) and recall ($0.268$ vs.\ $0.262$), the same PickScore and CLIP score, and a higher ImageReward ($0.706$ vs.\ $0.676$), at comparable precision. With $\theta_{\max} = 30^\circ$, it is better than MoNO on every quality metric, with an ImageReward of $0.738$ that is above even the student's $0.718$, while remaining more diverse than TWIST.

The oracle is also almost twice as fast as MoNO ($0.32$ vs.\ $0.58$\,s per image). Its loss is defined on the whole set, so all $32$ noises are generated, differentiated and updated together in batched passes. MoNO instead optimizes each noise to move away from the samples generated before it, so it must generate and optimize the samples one at a time.

When the two objectives are compared in the same setting, with the noises of each prompt optimized at inference under the same budget, the TWIST objective therefore matches or outperforms MoNO's: it reaches the same diversity at a higher quality, or a clearly higher quality at a moderate cost in diversity, in almost half the time. The lower recall of TWIST without selection is thus not a limitation of its objective but the price of amortization: TWIST trains a single network once and applies it to unseen prompts with no per-prompt optimization and no backward pass at inference.

% ------------------------------------------------------------------------------------------------ ReDiverse

\begin{table*}[t]
\centering
\caption{\textbf{Comparison with ReDiverse-FT}~\protect\cite{horoz2026rediverse} (128 test prompts, $n{=}32$, no selection). ReDiverse-FT uses the authors' code and SDXL recipe unchanged, trained on the same training prompts as TWIST; the adapter is active in the student's first step. All other settings are shared with TWIST and \cref{tab:main}.}
\label{tab:rediverse}
\small
\setlength{\tabcolsep}{4pt}

\begin{tabular}{ll ccc cccc}
\toprule
& & \multicolumn{3}{c}{Diversity} & \multicolumn{4}{c}{Quality / alignment} \\
\cmidrule(lr){3-5}\cmidrule(lr){6-9}
Model & Method & Rec.$\uparrow$ & Vendi$\uparrow$ & qVendi$\uparrow$ & Prec.$\uparrow$ & Pick$\uparrow$ & IR$\uparrow$ & CLIP$\uparrow$ \\
\midrule
\multirow{3}{*}{\rotatebox{90}{\scriptsize Turbo}}
 & Student & 0.114 & 3.30 & 2.12 & 0.577 & 22.39 & 0.718 & 26.28 \\
 & ReDiverse-FT & 0.187 & 5.07 & 2.64 & 0.307 & 21.64 & 0.227 & 25.56 \\
 & \cellcolor{oursbg}TWIST & \cellcolor{oursbg}0.164 & \cellcolor{oursbg}3.74 & \cellcolor{oursbg}2.38 & \cellcolor{oursbg}0.554 & \cellcolor{oursbg}22.33 & \cellcolor{oursbg}0.690 & \cellcolor{oursbg}26.14 \\
\midrule
\multirow{3}{*}{\rotatebox{90}{\scriptsize DMD2}}
 & Student & 0.214 & 3.82 & 2.53 & 0.559 & 22.10 & 0.790 & 26.75 \\
 & ReDiverse-FT & 0.354 & 5.45 & 3.08 & 0.391 & 21.67 & 0.432 & 26.20 \\
 & \cellcolor{oursbg}TWIST & \cellcolor{oursbg}0.332 & \cellcolor{oursbg}4.79 & \cellcolor{oursbg}2.99 & \cellcolor{oursbg}0.481 & \cellcolor{oursbg}21.89 & \cellcolor{oursbg}0.668 & \cellcolor{oursbg}26.61 \\
\midrule
\multirow{3}{*}{\rotatebox{90}{\scriptsize\shortstack{SDXL-\\Lightning}}}
 & Student & 0.454 & 4.72 & 2.99 & 0.625 & 22.55 & 0.697 & 26.07 \\
 & ReDiverse-FT & 0.529 & 5.11 & 3.14 & 0.619 & 22.48 & 0.630 & 25.92 \\
 & \cellcolor{oursbg}TWIST & \cellcolor{oursbg}0.538 & \cellcolor{oursbg}5.13 & \cellcolor{oursbg}3.19 & \cellcolor{oursbg}0.599 & \cellcolor{oursbg}22.46 & \cellcolor{oursbg}0.648 & \cellcolor{oursbg}26.01 \\
\bottomrule
\end{tabular}
\end{table*}

\subsection{Comparison with ReDiverse}
\label{app:rediverse}

ReDiverse~\cite{horoz2026rediverse} is concurrent work: it was released on November 13, 2026, after most of this work was done, so we compare with it here rather than in the main paper. It also restores the diversity of a distilled student with the help of its teacher, but at the level of individual activations. A probe first locates the block $\ell$ of the student's first step at which its activations collapse into fewer directions than the teacher's, together with the step $\tau$ of the teacher's trajectory whose activations serve as the target. ReDiverse-FT then trains a rank-$32$ LoRA on the blocks up to $\ell$ that regresses the student's activation at $\ell$ onto the teacher's activation at step $\tau$, for the same prompt and the same noise; the adapter is active in the first step only.

\paragraph{Setup.} We use the authors' code and their released recipe without changes: the collapsed block (\texttt{up\_blocks.0.attentions.0}) and target step ($\tau = 7$ of the $50$-step teacher trajectory) reported for SDXL, the LoRA targets, rank and optimizer, and $200$ steps of $4$ captions with $8$ noises each. We apply it to the same three students as TWIST: for the one-step SDXL-Turbo and DMD2, the adapter acts on the student's only forward pass, and for SDXL-Lightning on the first of its four. Everything else is shared with TWIST: the training prompts, the test prompts and noises, and the evaluation. The teacher targets of ReDiverse-FT are produced by its own recipe, from its own seeded noises and the teacher's trajectory up to step $\tau$. Both methods are amortized and add no backward pass at inference, so they are compared without selection.

\paragraph{Results.} \Cref{tab:rediverse} shows that, trained on the same data, TWIST gives a better balance between diversity and quality on every student. On SDXL-Lightning, TWIST is ahead in recall ($0.538$ vs.\ $0.529$), Vendi, qVendi ($3.19$ vs.\ $3.14$), ImageReward and CLIP score, and the two are close in precision and PickScore. On SDXL-Turbo and DMD2, ReDiverse reaches a slightly higher recall ($0.187$ vs.\ $0.164$ and $0.354$ vs.\ $0.332$) and a higher Vendi score, but at a large cost in quality: its precision drops to $0.307$ and $0.391$ (TWIST: $0.554$ and $0.481$) and its ImageReward to $0.227$ and $0.432$ (TWIST: $0.690$ and $0.668$), far below the student's. Its higher Vendi score on these students therefore reflects samples that leave the teacher's distribution rather than coverage of it.

\paragraph{Discussion.} The two methods differ in what they ask the student to match. ReDiverse matches activations one sample at a time: each student activation is regressed onto the teacher's activation for the same noise. This requires a pairing between the two, which is only meaningful if the teacher's early trajectory and the student's first step map the same noise to the same layout, and it requires the probe to find, for every teacher--student pair, the block $\ell$ and the teacher step $\tau$ at which this correspondence holds; the values released for SDXL were found for the four-step DMD2, and a different student may need a different pair. TWIST instead matches the shape of the set of samples, its spectrum and principal directions in feature space, and that loss does not depend on which student sample corresponds to which teacher sample (Proposition~\ref{prop:perm}, \cref{tab:supp_pair}). It needs no probe of the student's internals, works on the final images of any student without choosing a block or a teacher step, and still trains when the noises are unpaired.

% ------------------------------------------------------------------------------------------------ feature space of the metrics
\begin{table*}[t]
\centering
\caption{\textbf{Diversity in other feature spaces} (128 test prompts, $n{=}32$). Vendi and qVendi computed on DINOv2-L CLS features (as in the main paper), CLIP ViT-L/14 image embeddings and DreamSim (ensemble) embeddings; qVendi multiplies the Vendi score of each space by the mean sigmoid of the ImageReward of the samples. TWIST + sel.\ uses the largest pool of \cref{tab:ablation_sel} ($\alpha{=}0.1$), at a cost equal to or lower than MoNO $K{=}1$; ReDiverse-FT as in \cref{tab:rediverse}.}
\label{tab:metric_space}
\small
\setlength{\tabcolsep}{4pt}

\begin{tabular}{@{\hspace{4pt}}c@{\hspace{10pt}}l cc cc cc c}
\toprule
& & \multicolumn{2}{c}{DINOv2-L} & \multicolumn{2}{c}{CLIP ViT-L/14} & \multicolumn{2}{c}{DreamSim} & \\
\cmidrule(lr){3-4}\cmidrule(lr){5-6}\cmidrule(lr){7-8}
 & Method & Vendi$\uparrow$ & qVendi$\uparrow$ & Vendi$\uparrow$ & qVendi$\uparrow$ & Vendi$\uparrow$ & qVendi$\uparrow$ & IR$\uparrow$ \\
\midrule
\multicolumn{2}{l}{\textcolor{gray}{Teacher (reference)}} & \textcolor{gray}{5.62} & \textcolor{gray}{3.49} & \textcolor{gray}{2.36} & \textcolor{gray}{1.50} & \textcolor{gray}{5.39} & \textcolor{gray}{3.39} & \textcolor{gray}{0.662} \\
\midrule
 & Student & 3.30 & 2.12 & 1.72 & 1.13 & 3.03 & 1.97 & 0.718 \\
 & MoNO $K{=}1$ & 4.39 & 2.76 & 1.95 & 1.25 & 4.24 & 2.70 & 0.676 \\
 & ReDiverse-FT & 5.07 & 2.64 & 2.17 & 1.17 & 3.54 & 1.89 & 0.227 \\
 & \cellcolor{oursbg}TWIST & \cellcolor{oursbg}3.74 & \cellcolor{oursbg}2.38 & \cellcolor{oursbg}1.81 & \cellcolor{oursbg}1.18 & \cellcolor{oursbg}3.46 & \cellcolor{oursbg}2.23 & \cellcolor{oursbg}0.690 \\
\multirow{-5}{*}{\rotatebox{90}{\footnotesize Turbo}} & \cellcolor{oursbg}TWIST + sel.\ (512) & \cellcolor{oursbg}5.53 & \cellcolor{oursbg}3.56 & \cellcolor{oursbg}2.00 & \cellcolor{oursbg}1.31 & \cellcolor{oursbg}4.15 & \cellcolor{oursbg}2.69 & \cellcolor{oursbg}0.723 \\
\midrule
 & Student & 3.82 & 2.53 & 1.92 & 1.29 & 3.77 & 2.50 & 0.790 \\
 & MoNO $K{=}1$ & 5.26 & 3.28 & 2.17 & 1.38 & 5.41 & 3.41 & 0.648 \\
 & ReDiverse-FT & 5.45 & 3.08 & 2.37 & 1.39 & 4.55 & 2.61 & 0.432 \\
 & \cellcolor{oursbg}TWIST & \cellcolor{oursbg}4.79 & \cellcolor{oursbg}2.99 & \cellcolor{oursbg}2.13 & \cellcolor{oursbg}1.36 & \cellcolor{oursbg}4.76 & \cellcolor{oursbg}3.01 & \cellcolor{oursbg}0.668 \\
\multirow{-5}{*}{\rotatebox{90}{\footnotesize DMD2}} & \cellcolor{oursbg}TWIST + sel.\ (288) & \cellcolor{oursbg}6.71 & \cellcolor{oursbg}4.23 & \cellcolor{oursbg}2.30 & \cellcolor{oursbg}1.48 & \cellcolor{oursbg}5.44 & \cellcolor{oursbg}3.44 & \cellcolor{oursbg}0.696 \\
\midrule
 & Student & 4.72 & 2.99 & 2.09 & 1.35 & 4.24 & 2.72 & 0.697 \\
 & MoNO $K{=}1$ & 8.78 & 4.38 & 3.14 & 1.59 & 8.82 & 4.41 & 0.056 \\
 & ReDiverse-FT & 5.11 & 3.14 & 2.20 & 1.38 & 4.35 & 2.71 & 0.630 \\
 & \cellcolor{oursbg}TWIST & \cellcolor{oursbg}5.13 & \cellcolor{oursbg}3.19 & \cellcolor{oursbg}2.19 & \cellcolor{oursbg}1.40 & \cellcolor{oursbg}4.68 & \cellcolor{oursbg}2.94 & \cellcolor{oursbg}0.650 \\
\multirow{-5}{*}{\rotatebox{90}{\scriptsize\shortstack{SDXL-\\Lightning}}} & \cellcolor{oursbg}TWIST + sel.\ (224) & \cellcolor{oursbg}7.54 & \cellcolor{oursbg}4.76 & \cellcolor{oursbg}2.45 & \cellcolor{oursbg}1.57 & \cellcolor{oursbg}5.61 & \cellcolor{oursbg}3.55 & \cellcolor{oursbg}0.712 \\
\bottomrule
\end{tabular}
\end{table*}

\subsection{Diversity in other feature spaces}
\label{app:metric_space}

Recall, precision, Vendi and qVendi in the main paper are computed on DINOv2-L CLS features. Since TWIST is trained on DINOv2-B patch features and our selection kernel uses DINOv2-B features, we also measure diversity in two feature spaces unrelated to DINOv2: CLIP ViT-L/14 image embeddings~\cite{radford2021clip} and DreamSim~\cite{fu2023dreamsim}, an embedding tuned to human judgments of image similarity. In each space, Vendi uses the cosine kernel of the $32$ samples of a prompt, and qVendi multiplies it by the mean quality of the samples, the sigmoid of their ImageReward, as in the main paper. \Cref{tab:metric_space} reports the student, MoNO with $K = 1$, whose cost is equal to or higher than that of TWIST with selection, ReDiverse-FT, TWIST and TWIST with selection on all three students.

The gains of TWIST over the student hold in every feature space. Without selection, TWIST raises the CLIP Vendi from $1.72$ to $1.81$ on SDXL-Turbo, from $1.92$ to $2.13$ on DMD2 and from $2.09$ to $2.19$ on SDXL-Lightning, and the DreamSim Vendi from $3.03$ to $3.46$, $3.77$ to $4.76$ and $4.24$ to $4.68$, and selection raises both further. The diversity that TWIST adds is therefore not specific to the DINOv2 features it is trained and evaluated with.

The comparison with other methods depends more on the feature space. Against MoNO with $K = 1$, TWIST with selection keeps a higher qVendi in CLIP space on SDXL-Turbo and DMD2 ($1.31$ vs.\ $1.25$ and $1.48$ vs.\ $1.38$), matches it in DreamSim space on SDXL-Turbo ($2.69$ vs.\ $2.70$) and is slightly ahead on DMD2 ($3.44$ vs.\ $3.41$); on SDXL-Lightning, MoNO scores higher in both spaces, but at an ImageReward of $0.056$, a loss of quality that the sigmoid weighting of qVendi only partly reflects. The lead of TWIST with selection is largest in DINOv2 space, where its selection kernel is defined. ReDiverse-FT, whose DINOv2 Vendi on SDXL-Turbo is above TWIST's ($5.07$ vs.\ $3.74$), falls to a similar value in DreamSim space ($3.54$ vs.\ $3.46$), and TWIST has the higher DreamSim qVendi on all three students ($2.23$ vs.\ $1.89$, $3.01$ vs.\ $2.61$ and $2.94$ vs.\ $2.71$), consistent with the artifacts discussed in \cref{app:rediverse}.

% ------------------------------------------------------------------------------------------------ full-width tables

% ============================================================== Supp. D: qualitative results
% Figures in supp_qual/ (made by paper_figs/supp_qual.py, images recompressed with Ghostscript /printer).
% Labels used: tab:main, tab:rediverse, app:rediverse, sec:select.
\section{Qualitative results}
\label{app:qual}

For every method, we show the first samples of the $32$ generated for a prompt, and all methods except TWIST with selection start from the same noises, which are also the noises of the teacher's samples (pooled to $64{\times}64$ for SDXL-Turbo). TWIST with selection keeps $32$ of a larger pool ($512$ candidates for SDXL-Turbo, $288$ for DMD2 and $224$ for SDXL-Lightning, $\alpha = 0.1$, as in \cref{tab:main}); for it we show the first samples in the order of the greedy selection. All other methods use the settings of \Cref{tab:main}, and ReDiverse-FT those of \Cref{app:rediverse}.

\begin{figure*}[p]
\centering
\includegraphics[width=\textwidth]{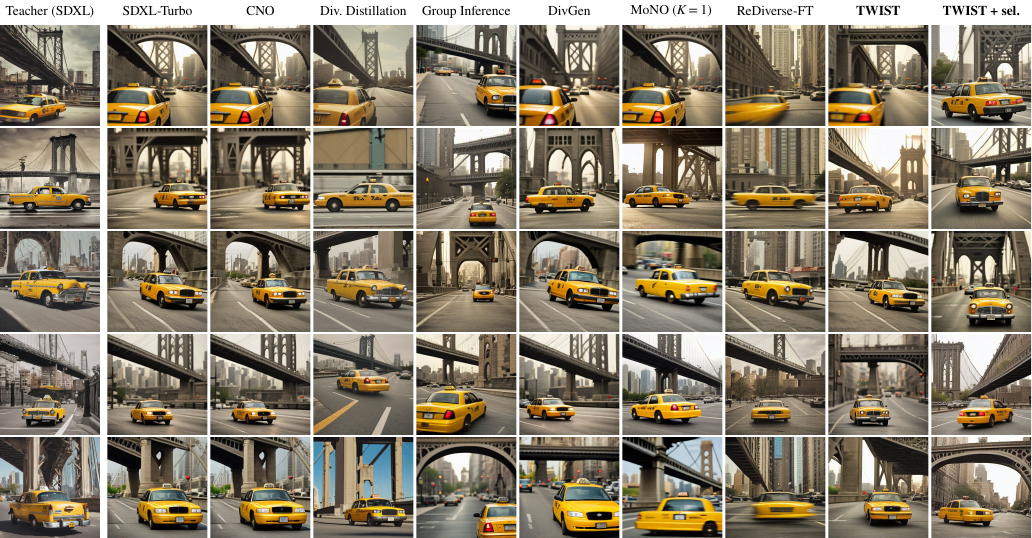}
\caption{\textbf{SDXL-Turbo}, ``A yellow taxi cab driving down a street near a bridge.'' First five of the $32$ samples of every method (uncurated; all methods except TWIST + sel.\ start from the same noises, see \cref{app:qual}).}
\label{fig:qualA_turbo_p272}
\vspace{1em}
\includegraphics[width=\textwidth]{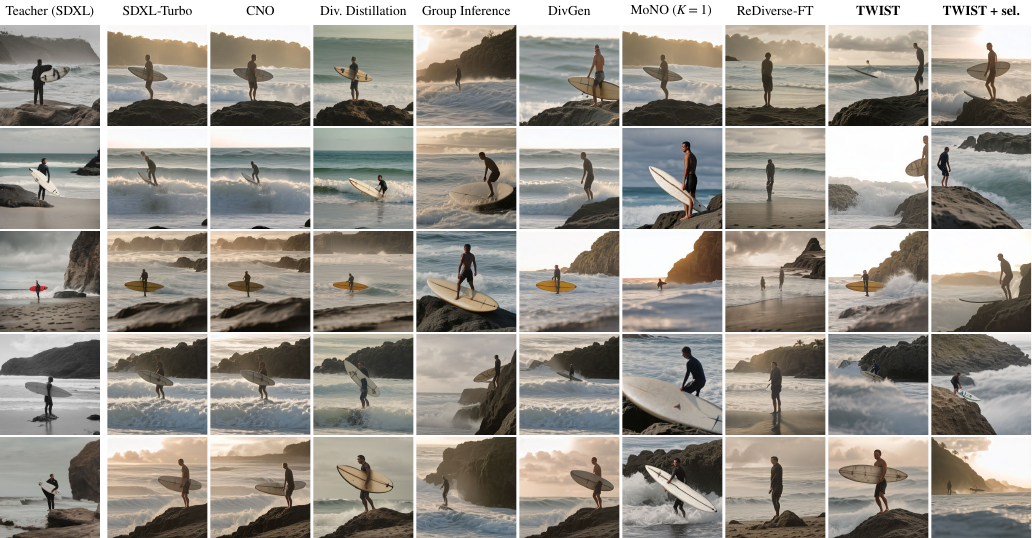}
\caption{\textbf{SDXL-Turbo}, ``A man standing on a rock holding a surfboard.'' First five of the $32$ samples of every method (uncurated; all methods except TWIST + sel.\ start from the same noises, see \cref{app:qual}).}
\label{fig:qualA_turbo_p322}
\end{figure*}

\begin{figure*}[p]
\centering
\includegraphics[width=\textwidth]{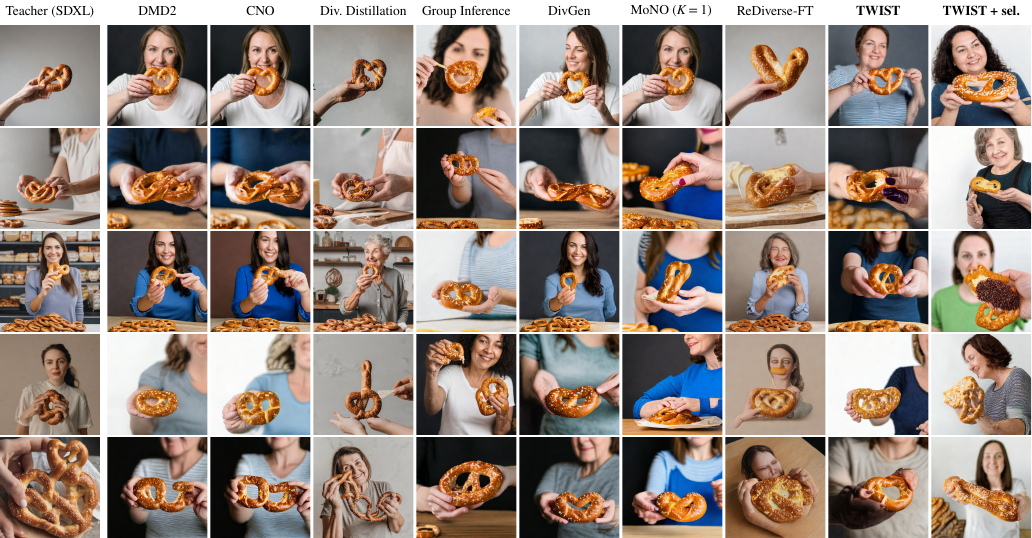}
\caption{\textbf{DMD2}, ``A lady is holding a pretzel with a cheese backing.'' First five of the $32$ samples of every method (uncurated; all methods except TWIST + sel.\ start from the same noises, see \cref{app:qual}).}
\label{fig:qualA_dmd2_p236}
\vspace{1em}
\includegraphics[width=\textwidth]{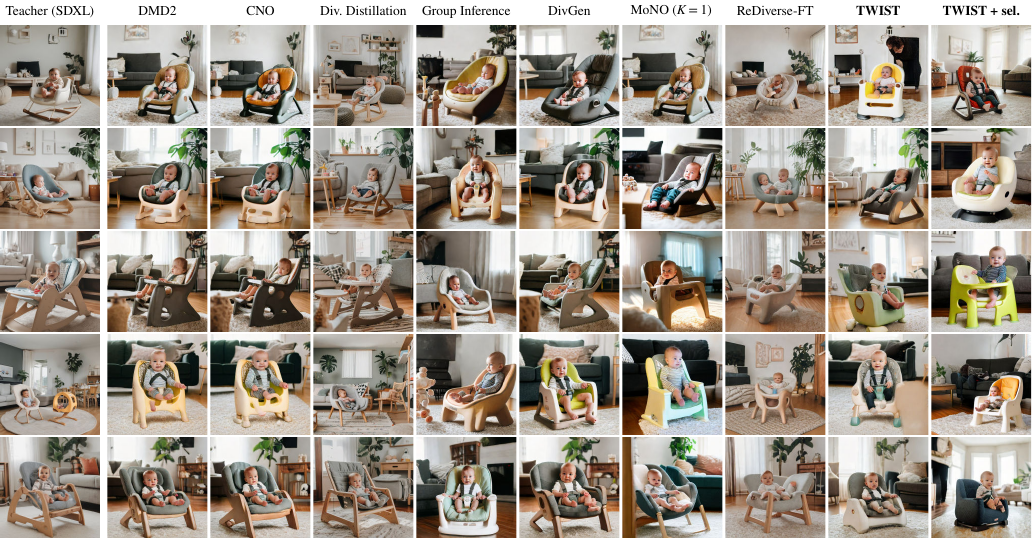}
\caption{\textbf{DMD2}, ``A baby in a play chair in a living room.'' First five of the $32$ samples of every method (uncurated; all methods except TWIST + sel.\ start from the same noises, see \cref{app:qual}).}
\label{fig:qualA_dmd2_p265}
\end{figure*}

\begin{figure*}[p]
\centering
\includegraphics[width=\textwidth]{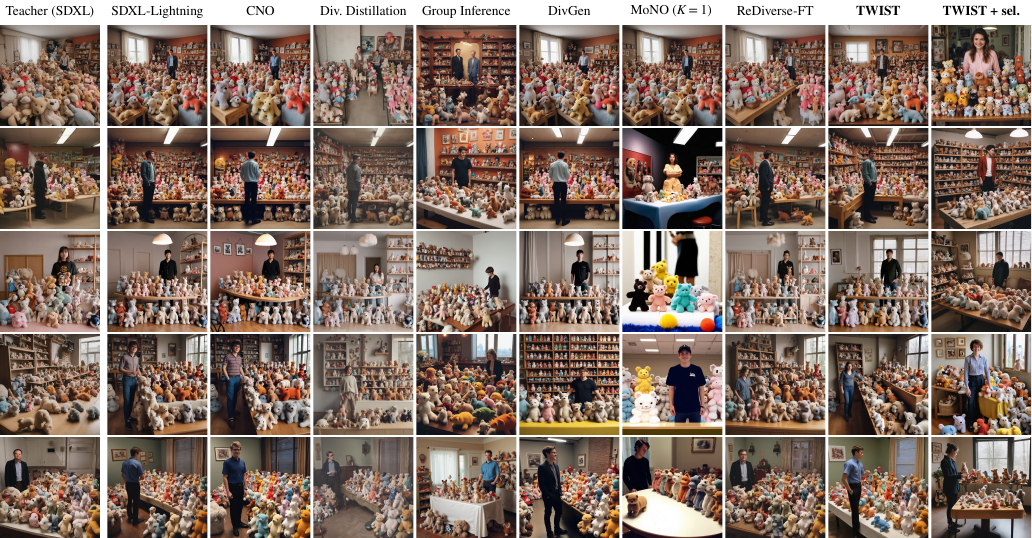}
\caption{\textbf{SDXL-Lightning}, ``A person standing next to a table with many stuffed animals.'' First five of the $32$ samples of every method (uncurated; all methods except TWIST + sel.\ start from the same noises, see \cref{app:qual}).}
\label{fig:qualA_light_p353}
\vspace{1em}
\includegraphics[width=\textwidth]{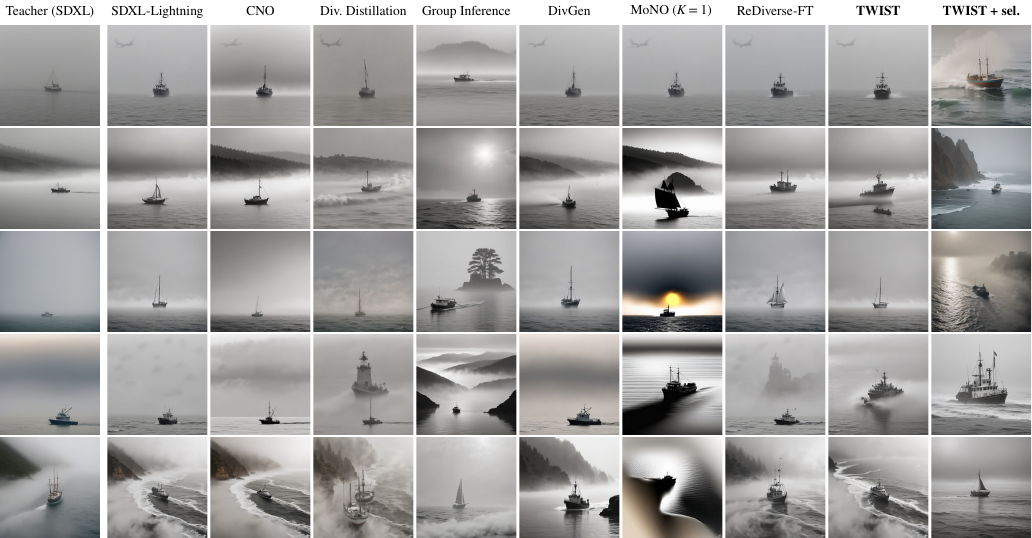}
\caption{\textbf{SDXL-Lightning}, ``A boat traverses the ocean in the midst of fog.'' First five of the $32$ samples of every method (uncurated; all methods except TWIST + sel.\ start from the same noises, see \cref{app:qual}).}
\label{fig:qualA_light_p308}
\end{figure*}

\paragraph{Comparison with all methods.}
\Cref{fig:qualA_turbo_p272,fig:qualA_turbo_p322,fig:qualA_dmd2_p236,fig:qualA_dmd2_p265,fig:qualA_light_p353,fig:qualA_light_p308} compare all methods on two prompts per student. Since the samples share their noises, CNO, which changes the student little, stays close to the student's images sample by sample, while Diversity Distillation, which runs the teacher's first step, changes layout and sometimes style at a cost in quality. TWIST alone changes the student's coarse layout for most noises and keeps it for some, and TWIST with selection varies layouts, viewpoints and subjects much more than the student while keeping its image quality.

\begin{figure*}[p]
\centering
\includegraphics[width=0.86\textwidth]{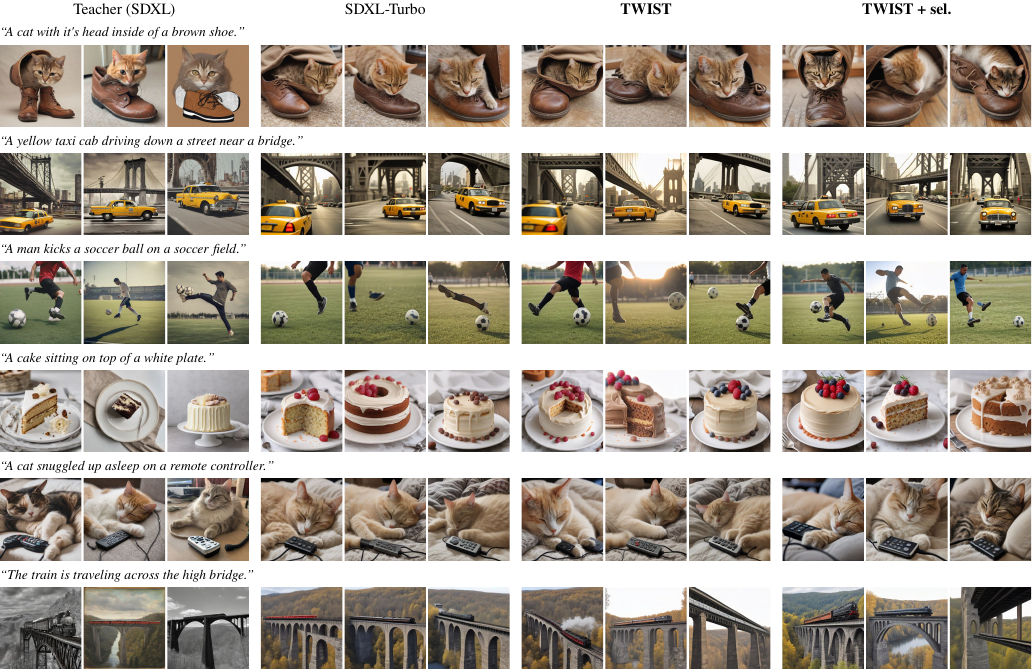}
\caption{\textbf{SDXL-Turbo} on six test prompts: the first three of the $32$ samples of the teacher, the student, TWIST and TWIST with selection (uncurated; all methods except TWIST + sel.\ start from the same noises, see \cref{app:qual}).}
\label{fig:qualB_turbo}
\vspace{1em}
\includegraphics[width=0.86\textwidth]{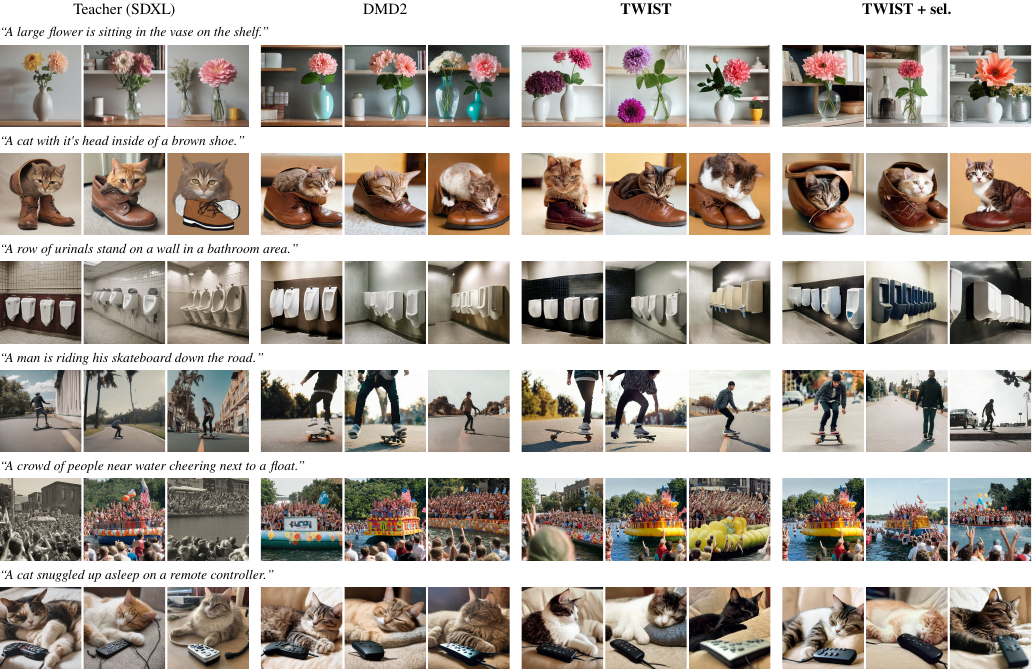}
\caption{\textbf{DMD2} on six test prompts: the first three of the $32$ samples of the teacher, the student, TWIST and TWIST with selection (uncurated; all methods except TWIST + sel.\ start from the same noises, see \cref{app:qual}).}
\label{fig:qualB_dmd2}
\end{figure*}

\begin{figure*}[p]
\centering
\includegraphics[width=0.86\textwidth]{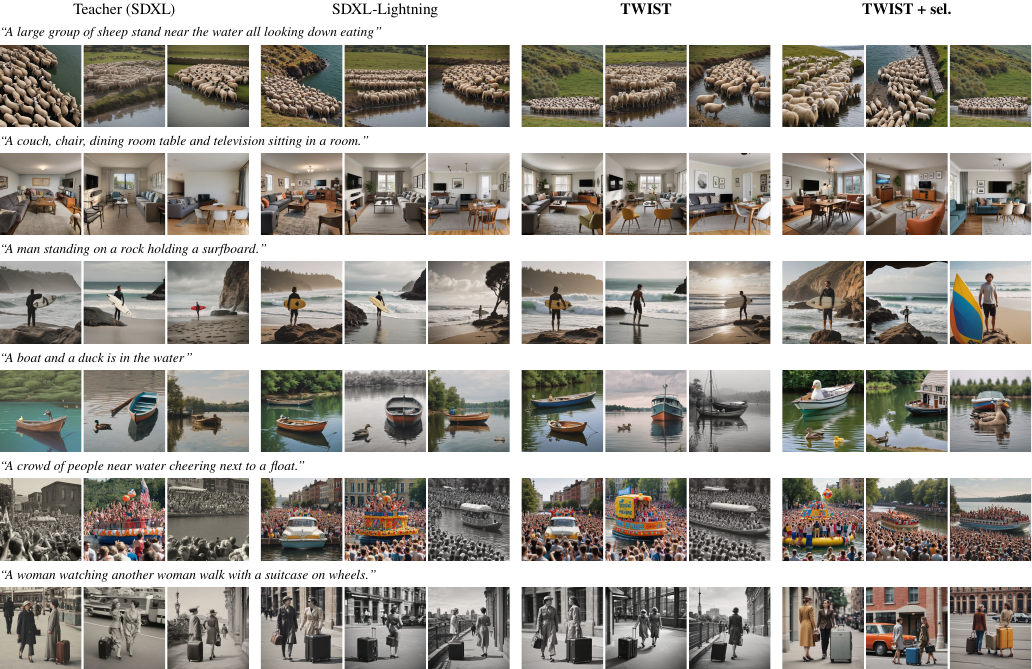}
\caption{\textbf{SDXL-Lightning} on six test prompts: the first three of the $32$ samples of the teacher, the student, TWIST and TWIST with selection (uncurated; all methods except TWIST + sel.\ start from the same noises, see \cref{app:qual}).}
\label{fig:qualB_light}
\vspace{1em}
\includegraphics[width=0.86\textwidth]{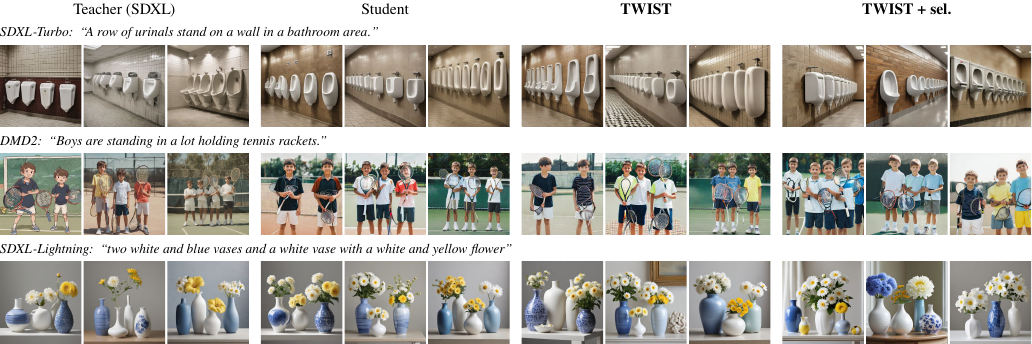}
\caption{\textbf{Failure cases.} For each student, the qualitative prompt on which TWIST loses the most recall relative to the student (first three samples, uncurated, as in \cref{app:qual}). The teacher varies the style or viewpoint of the image (drawn and photographic players, frontal and angled views), which TWIST does not recover.}
\label{fig:qualC}
\end{figure*}

\paragraph{Breadth across prompts.}
\Cref{fig:qualB_turbo,fig:qualB_dmd2,fig:qualB_light} show the teacher, the student, TWIST and TWIST with selection on six prompts per student, three samples each.

\paragraph{Failure cases.}
\Cref{fig:qualC} shows, for each student, the qualitative prompt on which TWIST has the largest drop in recall relative to the student. In these cases, the teacher's variation lies mostly in the style or viewpoint of the image, \eg drawn versus photographic tennis players, or frontal versus angled views of the urinals, while the student's samples share one style and viewpoint; TWIST varies the arrangement within them but does not recover the teacher's changes of style and viewpoint.

% ------------------------------------------------------------------------------------------------ A: all methods

% \clearpage
% \makeatletter\setcounter{NAT@ctr}{0}\makeatother
% {\small
% \bibliographystyleS{ieeenat_fullname}
% \bibliographyS{main,supp}
% }

\end{document}